\PassOptionsToPackage{dvipsnames,svgnames,table}{xcolor}
\documentclass[10pt,journal,compsoc]{IEEEtran}

\ifCLASSOPTIONcompsoc
  \usepackage[nocompress]{cite}
\else
  \usepackage{cite}
\fi

\ifCLASSINFOpdf
\else
\fi

\usepackage[utf8]{inputenc} 
\usepackage[T1]{fontenc}    
\usepackage{hyperref}       
\usepackage{url}            
\usepackage{booktabs}       
\usepackage{amsfonts}       
\usepackage{nicefrac}       
\usepackage{microtype}      
\usepackage{xcolor}         
\usepackage{graphicx}
\usepackage{subfigure}
\usepackage{amsmath,amssymb} 
\usepackage{caption} 

\usepackage{wrapfig}
\usepackage[american]{babel}

\usepackage{graphicx} 
\usepackage{times}
\usepackage{soul}
\usepackage{url}
\usepackage[utf8]{inputenc}
\usepackage{amsmath}
\usepackage{amsthm}
\usepackage{booktabs}
\usepackage{subfigure}
\usepackage{amsmath}
\usepackage{amssymb}
\usepackage{booktabs}
\usepackage{multirow}
\usepackage{paralist}
\usepackage[american]{babel}
\usepackage{microtype}
\usepackage{lipsum}
\usepackage{bm}
\usepackage[switch]{lineno}  %

\usepackage{subfigure}
\usepackage{amsmath,amssymb} 
\usepackage{caption} 
\usepackage{hyperref}
\usepackage{wrapfig}
\usepackage{subfigure}
\usepackage{csquotes}
\usepackage{bm}
\usepackage{paralist,algorithm}
\usepackage{multirow}

\usepackage[noend]{algpseudocode}

\newcommand{\x}{{\bf x}}
\newcommand{\eg}{{\em e.g.}}
\newcommand{\ie}{{\em i.e.}}

\newcommand{\p}{{\bf p}}

\newcommand{\w}{{\bf w}}

\newcommand{\ttt}{{\bf t}}

\newcommand{\z}{{\bf z}}

\newcommand{\D}{\mathcal{D}}

\newcommand{\R}{\mathbb{R}}

\newcommand{\name}{{\sc Proof }}
\newcommand{\mame}{{\sc Proof}}
\newcommand{\bfname}[1]{{\bf #1}}

\usepackage{accents}
\newlength{\dhatheight}

\usepackage{xcolor,colortbl}
\definecolor{Gray}{gray}{0.85}
\definecolor{Redo}{rgb}{0.95,0.69,0.51}
\definecolor{LightCyan}{rgb}{0.88,1,1}
\usepackage{afterpage}
\usepackage{pifont}

\usepackage{threeparttable}
\usepackage{amssymb}
\definecolor{DarkGreen}{RGB}{1,50,32}
\definecolor{dr}{RGB}{47,85,151}
\definecolor{dreg}{RGB}{112,48,160}
\definecolor{dn}{RGB}{255, 117, 143}
\definecolor{pr}{RGB}{228,197,111}
\definecolor{kd}{RGB}{251,133,0}
\definecolor{mr}{RGB}{84,130,53}

\usepackage{csquotes}
\begin{document}

\title{Learning without Forgetting for  \\Vision-Language Models}

\author{Da-Wei Zhou, Yuanhan Zhang, Yan Wang, Jingyi Ning, Han-Jia Ye, De-Chuan Zhan, Ziwei Liu
\thanks{
	This work is partially supported by National Science and Technology Major Project (2022ZD0114805), NSFC (62476123, 62376118, 62006112, 62250069), Fundamental Research Funds for the Central Universities (2024300373,14380021), CCF-Tencent Rhino-Bird Open Research Fund RAGR20240101, Collaborative Innovation Center of Novel Software Technology and Industrialization, China Scholarship Council, the AI \& AI for Science Project of Nanjing University, Ministry of Education, Singapore, under its MOE AcRF Tier 2 (MOET2EP20221- 0012), NTU NAP, and under the RIE2020 Industry Alignment Fund – Industry Collaboration Projects (IAF-ICP) Funding Initiative. 	(Corresponding authors: H.-J. Ye and Z. Liu.)

D.-W. Zhou, Y. Wang, J. Ning, H.-J. Ye,  and D.-C. Zhan are with School of Artificial Intelligence,  Nanjing University, and National Key Laboratory for Novel Software Technology, Nanjing University,  Nanjing, 210023, China;
E-mail: \{zhoudw, wangy, yehj, zhandc\}@lamda.nju.edu.cn, ningjy@smail.nju.edu.cn

	Y. Zhang and Z. Liu are with S-Lab, College of Computing and Data Science, Nanyang Technological University, Singapore, 639798. E-mail: yuanhan002@e.ntu.edu.sg, ziwei.liu@ntu.edu.sg

}}

%
%

\markboth{Journal of \LaTeX\ Class Files,~Vol.~14, No.~8, August~2015}%
{Shell \MakeLowercase{\textit{et al.}}: Bare Demo of IEEEtran.cls for Computer Society Journals}
%



\IEEEtitleabstractindextext{%
\begin{abstract}
	  Class-Incremental Learning (CIL) or continual learning is a desired capability in the real world, which requires a learning system to adapt to new tasks without forgetting former ones. While traditional CIL methods focus on {\em visual} information to grasp core features, recent advances in Vision-Language Models (VLM) have shown promising capabilities in learning generalizable representations with the aid of {\em textual} information. However, when continually trained with new classes, VLMs often suffer from catastrophic forgetting of former knowledge. Applying VLMs to CIL poses two major challenges: \textbf{1)} how to adapt the model without forgetting; and \textbf{2)} how to make full use of the multi-modal information. To this end, we propose PROjectiOn Fusion (\textbf{\textsc{Proof}}) that enables VLMs to learn without forgetting. To handle the first challenge, we propose training task-specific projections based on the frozen image/text encoders. When facing new tasks, new projections are expanded, and former projections are fixed, alleviating the forgetting of old concepts. For the second challenge, we propose the fusion module to better utilize the cross-modality information. By jointly adjusting visual and textual features, the model can capture better task-specific semantic information that facilitates recognition.  Extensive experiments on nine benchmark datasets with various continual learning scenarios and various VLMs validate that \name achieves state-of-the-art performance. Code is available at \url{https://github.com/zhoudw-zdw/PROOF}.

\end{abstract}

}

\maketitle

\IEEEdisplaynontitleabstractindextext

%
\IEEEpeerreviewmaketitle



\section{Introduction}

In our ever-changing world, training data often comes in a stream format with new classes, requiring a learning system to absorb them continually~\cite{gomes2017survey,chao2020revisiting,geng2020recent,liu2024survey,ye2020heterogeneous}. To address the challenge of learning emerging new classes, Class-Incremental Learning (CIL) has been proposed~\cite{rebuffi2017icarl}. However, in CIL, the absence of former classes triggers catastrophic forgetting~\cite{french1999catastrophic}, where learning new concepts overwrites the knowledge of old ones and results in a decline in performance~\cite{li2016learning}. Numerous efforts have been made~\cite{de2021survey,masana2022class} to combat catastrophic forgetting in the machine learning field.

With the rapid development of pre-training techniques~\cite{han2021pre,wen2023large,yang2023exploring,wang2024survey}, recent years have witnessed the transition of CIL research from training from scratch~\cite{wu2019large,hou2018lifelong,zhao2020maintaining} to {\em utilizing pre-trained models} (PTM)~\cite{wang2022dualprompt,wang2022learning,seale2022coda,zhou2024continual}. With the help of PTM, \eg, Vision Transformers~\cite{dosovitskiy2020image}, incremental learners are born with strong transferability to grasp the {\em visual} features.
Facing the domain gap introduced by the incremental classes, they only need to learn a limited number of additional parameters~\cite{jia2022visual,chenadaptformer,lianscaling} as the {\em patches} to bridge the distribution gap, which significantly simplifies the challenge of incremental learning.

While pre-trained ViT-based CIL methods focus on learning the {\em visual} features to recognize new concepts, recent advances in Vision-Language Models (VLM) have demonstrated the potential of {\em textual} information in building generalized feature representations. A seminal work, \ie, contrastive language-image pre-training~\cite{radford2021learning} (CLIP), maps the visual and textual information in the shared embedding space, enabling robust learning and recognition of concepts from diverse sources. 
This integration of visual and textual modalities presents a promising avenue for developing continual learning models that can effectively adapt to real-world scenarios. 

Extending VLMs to CIL faces two significant challenges.
Firstly, sequentially tuning the VLM overwrites the innate generalizability and former concepts,  with the former leading to poor performance on future tasks and the latter to catastrophic forgetting.  Secondly, relying solely on textual information for classification neglects the valuable cross-modal features present in the multi-modal inputs. To fully utilize this information, it is necessary to explore methods for cross-modal fusion beyond textual features.

Correspondingly, we aim to turn a VLM into a continual learner that is both {\em retentive} and {\em comprehensive} so that VLMs can be updated in an incremental manner.  
Retentive refers to the model's ability to maintain its pre-trained capabilities, thereby preserving generalizability and enabling it to perform well on future tasks without forgetting.
Comprehensive refers to the model's capacity to integrate and adjust information from multiple modalities.
By leveraging these characteristics, we can mitigate catastrophic forgetting and use cross-modal features to build more robust classifiers as data evolves.

In this paper, we propose PROjectiOn Fusion (\textbf{\textsc{Proof}}) to address catastrophic forgetting in VLM. To make the model retentive, we freeze the pre-trained image/text backbones and append liner projections on top of them.
The task-specific information is encoded in the corresponding projection layer by mapping the projected features.
 When facing new tasks, new projections are extended while old ones are frozen, preserving former knowledge. 
 Besides, we aim to fuse the information from different modalities via cross-modal fusion, which allows for the query embedding to be adjusted with context information. Consequently, \name efficiently incorporates new classes and meanwhile resists forgetting old ones, achieving state-of-the-art performance on nine benchmark datasets and a non-overlapping TV series classification dataset. 
 We also evaluate \name in various continual learning settings, including CIL and continual cross-modal retrieval, to show its effectiveness in various real-world scenarios. Our contributions can be summarized as follows:

\begin{itemize}
	\item We propose a general framework that enables a pre-trained vision-language model to continually learn new classes without catastrophic forgetting;
	\item We design a novel projection fusion mechanism to enhance the model's representation ability and a cross-modal fusion module to encode task-specific information. We build the aggregated inference format considering cross-modal matching targets holistically;
	\item \name achieves state-of-the-art performance on nine benchmark datasets and a non-overlapping  dataset. Benefiting from its universality, \name also shows strong performance on continual cross-modal retrieval tasks against other cutting-edge methods.
\end{itemize}


\section{Related Work}

\subsection{Vision-Language Model (VLM) Tuning}
Recent years have witnessed the prosperity of research in vision-language models, \eg, CLIP~\cite{radford2021learning}, ALIGN~\cite{jia2021scaling}, CoCa~\cite{yu2022coca}, Florence~\cite{yuan2021florence}, BLIP~\cite{li2023blip}, CLIPPO~\cite{tschannen2022image}, and Flamingo~\cite{alayrac2022flamingo}. 
These models are pre-trained on vast amounts of images and texts, achieving a unified embedding space across different modalities. 
With great generalizability, they can be applied for downstream tasks in a zero-shot manner. 
However, a domain gap still exists between the pre-trained and downstream datasets, requiring further tuning for better performance. 
To fill this gap, many methods are proposed to tune a pre-trained VLM for downstream tasks.
CoOp~\cite{zhou2022learning} applies prompt learning~\cite{li2021prefix} into VLM tuning with learnable prompt tokens for the textual branch, where a set of learnable prompts are utilized to replace the template texts.  CoCoOp~\cite{zhou2022conditional} further encodes the instance-specific visual features into the learnable prompts.
CLIP-Adapter~\cite{gao2021clip} appends linear adapters after the visual and textual encoders to align the embeddings in the adapted space. ProDA~\cite{lu2022prompt} introduces prompt distribution learning into the prompt learning process, and TaskRes~\cite{yu2022task} directly learns a task-wise residual feature to bridge the domain gap. 
Tip-Adapter~\cite{zhang2022tip} caches visual prototypes and combines them with textual encoded information to construct a cross-modal inference paradigm.
Moreover, PLOT~\cite{chen2023plot} optimizes cross-modal matching by aligning multiple local visual features with textual prompts via optimal transport~\cite{monge1781memoire}. 
\cite{lee2023multimodal} proposes a prompt learning technique to tackle the missing modality with a pre-trained multi-modal transformer.
 Recent works also utilize ChatGPT as auxiliary knowledge to enhance the cross-modal mapping process~\cite{mao2022doubly,menon2022visual}.
 However, these works only focus on adapting VLMs to downstream tasks while overlooking the {\em catastrophic forgetting} of previous ones. When deploying VLMs into a sequence of downstream tasks, a desired algorithm should handle all tasks without forgetting.

 \subsection{Class-Incremental Learning (CIL)}
Class-Incremental Learning aims to learn from evolutive data and absorb new knowledge without forgetting~\cite{zhou2023class,masana2022class,de2021survey,zhou2024continual,ye2024contextualizing}, which can be divided into several groups. Replay-based methods~\cite{luo2023class,aljundi2019gradient,chaudhry2018riemannian,liu2020mnemonics,chaudhry2018efficient} save and replay former instances (\ie, exemplars) to recover old knowledge when learning new ones. Apart from directly replaying raw images, there are also works considering replaying features~\cite{iscen2020memory}, low-resolution images~\cite{zhao2021memory}, and logits~\cite{buzzega2020dark}. Moreover, generative models are also widely applied to model the distribution of previous tasks for replay, \eg, GAN~\cite{ostapenko2019learning,xiang2019incremental}, VAE~\cite{sun2022exemplar}, diffusion model~\cite{jodelet2023class,gao2023ddgr}. The second group utilizes knowledge distillation~\cite{hinton2015distilling} to build the mapping between models as regularization term~\cite{rebuffi2017icarl,li2016learning,douillard2020podnet}.
iCaRL~\cite{rebuffi2017icarl} and LwF~\cite{li2016learning} explore the logit-wise distillation to align the predictions between old and new models to resist forgetting. Besides, \cite{hou2019learning,lu2022augmented,park2021class} build the feature-wise mapping, and \cite{gao2022rdfcil,tao2020topology,dong2021few} regularize the group-wise information via relational distillation. The third group builds parameter-wise regularization terms to force important parameters not to drift away~\cite{kirkpatrick2017overcoming,aljundi2018memory,zenke2017continual,aljundi2019task}. The fourth group locates and rectifies the inductive bias of CIL models for unbiased predictions~\cite{shi2022mimicking,zhao2020maintaining,wu2019large,yu2020semantic}. For example, BiC~\cite{wu2019large} finds the predicted logits of the latest task are much higher than previous ones and designs a bias correction layer to calibrate the prediction. 
IL2M~\cite{belouadah2019il2m} calibrates the logits via re-scaling the task-wise predictions, while WA~\cite{zhao2020maintaining} directly normalizes the fully-connected layers for an unbiased prediction. The last group designs dynamic networks~\cite{yan2021dynamically,wang2022foster,zhou2022model} by expanding the network structure as data evolves. The network expansion techniques are further divided into neuron-wise~\cite{yoon2018lifelong,xu2018reinforced}, backbone-wise~\cite{yan2021dynamically,wang2022foster,zhou2022model,wang2023beef}, and token-wise~\cite{douillard2022dytox}. Besides, OSN~\cite{hutask} contains shared knowledge induced network partition and sharpness-aware orthogonal sparse network learning, aiming to enhance the plasticity and capacity.

 \subsection{CIL with VLM}
The aforementioned CIL algorithms aim to train an incremental model from scratch, while it would be easier to start with a pre-trained model~\cite{lee2023pre}. The integration of pre-trained Vision Transformer~\cite{dosovitskiy2020image} into CIL has attracted the attention of the community, and most methods~\cite{wang2022dualprompt,wang2022learning,seale2022coda} employ parameter-efficient tuning techniques to learn without forgetting.
L2P~\cite{wang2022learning} introduces the prompt pool and prompt search mechanism in CIL. It freezes the pre-trained weights, optimizes a set of visual prompts, and searches for the most similar prompts for instance-specific embeddings. DualPrompt~\cite{wang2022dualprompt} further explores the prompt depth and shared prompt for all tasks. CODA-Prompt~\cite{seale2022coda} replaces the key-value search mechanism and designs an attention-based prompt calculation strategy. NSP$^2$~\cite{lu2024visual} aims to learn each task by tuning the prompts in the direction orthogonal to the subspace spanned by previous tasks’ features, so as to ensure no interference on tasks that have been learned to overcome catastrophic forgetting.
Following works also consider generating prompts via a meta-network~\cite{jung2023generating,tang2023prompt} or aggregating predictions via a set of adjusted models~\cite{Gao_2023_ICCV,wang2022isolation}. However, these works are designed for pre-trained ViT and lack the potential to be compatible with vision-language models with multiple modality information. 
 S-Prompts~\cite{wang2022s} explores CLIP in {\em domain}-incremental learning, but the application of VLM in CIL remains relatively unexplored. WiSE-FT~\cite{wortsman2022robust} utilizes weight ensemble for robust finetuning, while it cannot be extended to multiple tasks.
This paper aims to address this research gap by presenting a comprehensive solution for tuning vision-language models without suffering from forgetting.

\section{Preliminaries}
In this section, we introduce the background information about class-incremental learning and vision language models. We also discuss the na\"ive solutions for tuning VLM in CIL.

\subsection{Class-Incremental Learning}
Given a data stream with emerging new classes, class-incremental learning aims to continually incorporate the knowledge and build a unified classifier~\cite{zhou2023class}. 
We denote the sequence of $B$ training sets without overlapping classes as $\left\{\D^{1}, \D^{2}, \cdots, \D^{B}\right\}$, where $\D^{b}=\left\{\left(\x_{i}, y_{i}\right)\right\}_{i=1}^{n_b}$ is the $b$-th training set with $n_b$ instances.
A training instance $\x_i \in \R^D$  belongs to class $y_i \in Y_b$. $Y_b$ is the label space of task $b$, and $Y_b  \cap Y_{b^\prime} = \varnothing$ for $b\neq b^\prime$. Following the typical CIL setting~\cite{rebuffi2017icarl,hou2019learning,wu2019large}, a fixed number of {\em exemplars} from the former classes are selected as the exemplar set $\mathcal{E}$. 
During the $b$-th incremental stage,  we can only access data from $\D^b$ and $\mathcal{E}$ for model training.
The target is to build a unified classifier for all seen classes $\mathcal{Y}_b=Y_1 \cup \cdots Y_b$ continually. In other words, we aim to find a model $f(\x): X\rightarrow\mathcal{Y}_b$ that minimizes the expected risk:
\begin{equation} \label{eq:totalrisk} 
	f^*=\underset{f \in \mathcal{H}}{\operatorname{argmin}} \; \mathbb{E}_{(\mathbf{x}, y) \sim \mathcal{D}_{t}^1\cup\cdots\mathcal{D}_{t}^b} \mathbb{I}\left(y \neq f(\mathbf{x})\right) \,,
\end{equation}
where $\mathcal{H}$ denotes the hypothesis space and $\mathbb{I}(\cdot)$ is the indicator function. $\mathcal{D}_{t}^b$ denotes the data distribution of task $b$. Following \cite{wang2022dualprompt,wang2022learning,wang2022s}, we assume that a pre-trained vision-language model is available as the initialization for $f(\x)$, which will be introduced in Section~\ref{sec:vlm}.

\subsection{Vision-Language Model} \label{sec:vlm}
This paper mainly focuses on contrastive
language-image pre-training (CLIP)~\cite{radford2021learning} as the VLM, while the proposed method is also compatible with other VLMs in Section~\ref{sec:exp:otherVLM}. During pre-training, CLIP jointly learns an image encoder $g_i(\cdot): \R^D\rightarrow\R^d$ and a text encoder $g_t(\cdot): \R^{Dt}\rightarrow\R^d$ in a contrastive manner, where $D/Dt$ are input dimensions of image/text, and $d$ is the embedding dimension.
CLIP projects a batch of image-text pairs into a shared embedding space. 
It maximizes the cosine similarity of paired inputs and minimizes it for unmatched ones. Benefiting from the massive training data, CLIP can synthesize a {\em zero-shot classifier} that generalizes to unseen classes. The output of CLIP is formulated as follows:
\begin{equation} \label{eq:clip_pred}
	p(y_i \mid \x)=\frac{\exp \left(\cos \left(\z, \w_i  \right) / \tau\right)}{\sum_{j=1}^{|\mathcal{Y}_b|} \exp \left(\cos \left(\z, \w_j \right) / \tau\right)} \,,
\end{equation}
where $\cos(\cdot,\cdot)$ denotes cosine similarity, $\tau$ is learnable temperature parameter, $\z=g_i(\x)$ is the image embedding. Correspondingly,  $\w_i$ is the text embedding of class $y_i$ obtained by feeding templated texts, \eg, ``a photo of a [CLASS]'' into the text encoder. 
We denote the templated text of class $i$ as $\mathbf{t}_i$. Eq.~\ref{eq:clip_pred} aims to find the most similar text $\mathbf{t}_i$ that maximizes the cosine similarity with the query image.

\subsection{Overcome Forgetting in Class-Incremental Learning} \label{sec:baseline}

Class-incremental learning, as a long-standing problem, has garnered significant attention from the research community.
In this section, we introduce two typical solutions for adapting pre-trained models with new classes and discuss their limitations.

\noindent\textbf{Vision-Based Learning:} 
Traditional CIL methods primarily rely on the {\em image encoder} to capture the patterns of new classes.
One such method, L2P~\cite{wang2022learning}, leverages visual prompt tuning~\cite{jia2022visual} 
to enable incremental updates of a pre-trained Vision Transformer~\cite{dosovitskiy2020image}.
By keeping the image encoder frozen, L2P trains a learnable prompt pool $\mathbf{Pool}$ and combines it with patch embeddings to obtain instance-specific embeddings. The optimization target can be formulated as:
\begin{equation} \label{eq:l2p} 
	\mathcal{L}= \ell \left(h
	\left( \bar{g_i}\left(\x_{i}, \mathbf{Pool} \right)	\right) , {y}_{i}\right)+ \mathcal{L}_{reg} \,,
\end{equation}
where $h(\cdot)$ is the classification head, $\bar{g_i}$ is the frozen image encoder, $\mathcal{L}_{reg}$ is the regularization loss for prompt selection. By freezing the encoder, Eq.~\ref{eq:l2p} grasps the new pattern with limited forgetting.

\noindent\textbf{CLIP Tuning:} 
The issue of tuning VLM without forgetting in CIL remains unaddressed, as previous works have solely focused on transferring CLIP to downstream tasks without considering the performance of former tasks. 
For instance, CoOp~\cite{zhou2022learning} converts  text template into a learnable prompt, \ie, $\mathbf{t}_i=$ [V]$_1$[V]$_2\cdots$[V]$_M$[CLASS]$_i$ and changes Eq.~\ref{eq:clip_pred} into:
\begin{equation} \label{eq:coop}
	p(y_i \mid \x)=\frac{\exp \left(\cos \left(\z, g_t(\ttt_i) \right) / \tau\right)}{\sum_{j=1}^{|\mathcal{Y}_b|} \exp \left(\cos \left(\z, g_t(\ttt_j) \right) / \tau\right)} \,.
\end{equation}
With the help of the learned prompt, Eq.~\ref{eq:coop} enables the model to be transferred to the downstream task. However, since the prompt template is shared for all tasks, sequentially tuning CoOp will suffer catastrophic forgetting of former concepts.

\noindent\textbf{Discussions:} 
Current methods focus on different aspects of CIL. Vision-based methods (\eg, Eq.~\ref{eq:l2p}) address the issue of forgetting but neglect the valuable semantic information conveyed in texts. 
Conversely, CLIP's pre-trained text encoder captures class-wise relationships that can enhance model learning.
Meanwhile, transfer learning methods (\eg, Eq.~\ref{eq:coop}) effectively leverage the cross-modal information while sequentially tuning them suffers the catastrophic forgetting of former concepts. 
Is it possible to combine the cross-modal information while resisting catastrophic forgetting?


\section{{\scshape{Proof}}: Projection Fusion for VLM}

Observing the limitations of typical vision-based methods in utilizing textual information and the forgetting phenomenon in CLIP tuning, we aim to leverage cross-modality knowledge in CLIP while effectively mitigating forgetting. 
To this end, we must make the model {\em retentive} and {\em comprehensive}. 
Retentive represents the ability to adapt to downstream tasks without forgetting, and we propose applying projections to map the pre-trained features into the projected feature space. 
Our unique training strategy ensures the preservation of former knowledge by freezing old projections and expanding new ones for new tasks. 
The comprehensive aspect involves co-adapting and utilizing cross-modal information to enhance unified predictions. 
The query instance's embedding is influenced by both visual and textual information, allowing for instance-specific adaptation and enabling comprehensive predictions.
In the following sections, we introduce the learning paradigm and the co-adaptation process. Lastly, we provide detailed guidelines for training and inference.

\begin{figure*}[t]
	\begin{center}
		{\includegraphics[width=2\columnwidth]{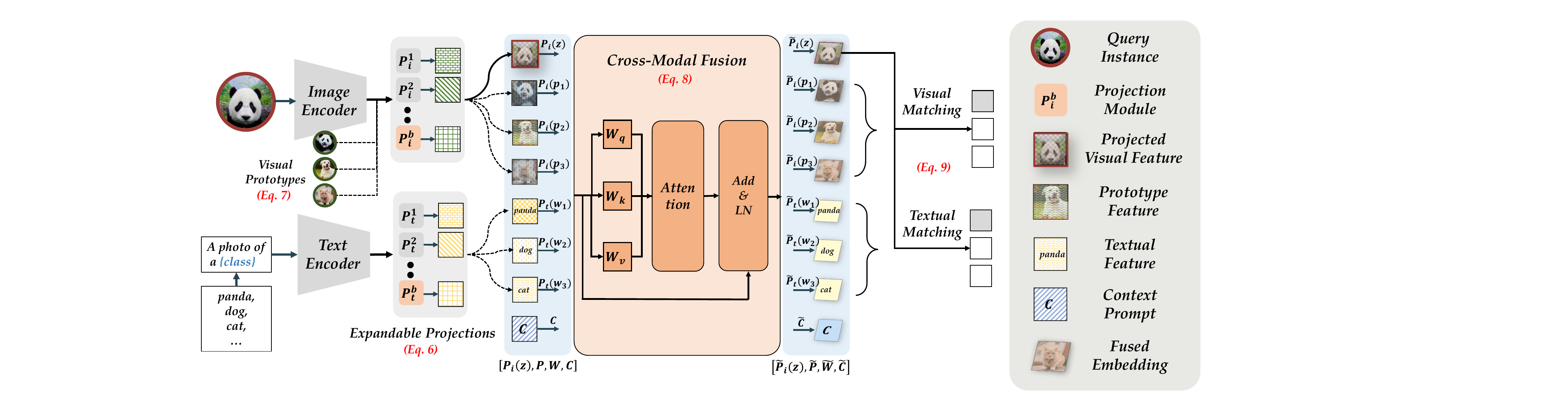}}
	\end{center}
	\caption{  Illustration of \mame. 
		The model learns expandable projections and aggregates them to get the aggregated features. The query instance, prototype features, textual features, and context prompts are fed into the cross-modal fusion module. 
		The fusion process utilizes self-attention to co-adapt the input set, which outputs the adapted features.
		The adapted query embedding is separately matched among the visual prototypes and textual features to get the final prediction. {\color{Redo}Red}  parts are trainable while {\color{gray}gray} ones are frozen.
	}
	\label{figure:teaser}
\end{figure*}

\subsection{Expandable Feature Projection}

CLIP is known for its strong zero-shot performance, \ie, Eq.~\ref{eq:clip_pred} obtains competitive results even without explicit training on the specific tasks. However, given the domain gap between pre-trained and downstream tasks, an {\em adaptation} process is still necessary to capture the characteristics of the latter. Specifically, we introduce a {\em linear layer} (denoted as ``{\bf projection}''), which is appended after the frozen image and text embeddings to facilitate the matching of pair-wise projected features. Denoting the projection of image/text as $P_i(\cdot):\R^d\rightarrow\R^d$ and $ P_t(\cdot):\R^d\rightarrow\R^d$, Eq.~\ref{eq:clip_pred} is transformed into:
\begin{equation} \label{eq:proj_pred}
	p(y_i \mid \x)=
	\underbrace{
	\frac{\exp \left(\cos \left( P_i \left(\z\right), P_t \left(\w_i\right)  \right) / \tau\right)}{\sum_{j=1}^{|\mathcal{Y}_b|} \exp \left(\cos \left( P_i \left(\z\right), P_t \left(\w_j\right) \right) / \tau\right)}
	}_\text{Projected Matching}
	 \,.
\end{equation}
We denote the classification based on Eq.~\ref{eq:proj_pred} as $f_\textbf{PM}(\x)$.
By freezing the image and text encoders, the downstream features in the projected space are aligned, allowing the model to encode the relevant downstream information into projection layers. 
Since the pre-trained model outputs generalizable features, the projection layer learns to {\em recombine} features in a data-driven manner. For instance, in a task involving `birds,' the projection would assign a higher weight to features like `beaks' and `wings.'
This adaptation enables the projected features to better discern and recognize downstream tasks.

\noindent\textbf{Expandable Projections:} However, sequentially training a {\em single} projection layer still leads to forgetting of former tasks, resulting in confusion when combining old and new concepts. To this end, we expand {\em task-specific} projections for each new task. Specifically, we append a newly initialized projection layer $P_i^b, P_t^b$ when a new task $\mathcal{D}^b$ arrives. This results in a set of projections: $\{P_i^1, P_i^2, \cdots P_i^b,\}$, $\{P_t^1, P_t^2, \cdots P_t^b,\}$, and we adopt the {\em aggregation} as the output:
\begin{equation} \label{eq:proj_sum} \textstyle
P_i(\z)=\sum_{m=1}^b P_i^m \left(\z\right), \quad  P_t(\w)=\sum_{n=1}^b P_t^n \left(\w\right)  \,.
\end{equation}
In Eq.~\ref{eq:proj_sum}, projected features from different stages are mapped and aggregated to capture the different emphases of former and latter tasks. 
For example, former tasks might emphasize `beak' features for bird recognition, while later tasks may focus on `beard' features to differentiate cats. 
The aggregation of different projections produces a comprehensive representation of the query instance.
 By substituting Eq.~\ref{eq:proj_sum} into Eq.~\ref{eq:proj_pred}, the model aligns the unified features in the joint space. Since Eq.~\ref{eq:proj_sum} only aggregates the projected features as the final representation, it has no restriction on the number of classes per task.

 \noindent\textbf{How to resist forgetting of former projections?} To overcome forgetting old concepts, we freeze the projections of former tasks when learning new ones, \ie, $\{\bar{P}_i^1, \bar{P}_i^2, \cdots P_i^b,\}$ (same for $P_t$). It allows the newly initialized projection to learn the {\em residual} information of new tasks,  incorporating new concepts while preserving the knowledge of former ones. 
 During the learning process of task $b$, we optimize the cross-entropy loss to encode the task-specific information into the current projections.

\noindent\textbf{Effect of projections}: The illustration of projections is shown in  Figure~\ref{figure:teaser} (left). \name learns projections based on the pre-trained encoders, which fits new patterns and maintains the generalizability of the pre-trained model. 
The parameter number of each projection layer is $d\times d$, which is negligible for the pre-trained model. These projections can be further merged during inference to alleviate the storage budget, as discussed in Section~\ref{sec:parameter}. 
Furthermore, the model learns new projections for new tasks, and task-specific projections fit new concepts easily. Since we only optimize the current projections and freeze old ones, the former knowledge is preserved, and forgetting is alleviated.

\subsection{Contextualizing Projections with Projection Fusion}
In Eq.~\ref{eq:proj_pred}, the projected visual and textual features are directly matched in the joint space. 
However, it would be beneficial to further {\em refine} these features to capture the {\em contextual relationship} between images and texts.
For instance, when the query instance is a `panda,' it is desirable to adjust the visual and textual features in a {\em coherent} manner to highlight discriminative attributes such as {\em black eyes and ears}. 
Similarly, when the query instance is a `cat,' features like beards and tails should be emphasized.
 This adjustment process involves jointly adapting the query embedding and the context (\eg, textual information) to obtain a {\em contextualized} embedding. Correspondingly, we propose a {\em set-to-set} function that contextualizes and fuses the query embeddings and contextual information.

Specifically, we denote the adaptation function as $\mathcal{T}(\cdot)$. It receives the query instance and context information as bags, \ie, $[P_i(\z), \textbf{Context}]$, and outputs the set of adjusted embeddings while being permutation-invariant: $\mathcal{T}([P_i(\z), \textbf{Context}])=[\tilde{P}_i(\z), \tilde{\textbf{Context}}]$. 
$\mathcal{T}(\cdot)$ encodes the set information and performs adaptation on each component.
 In the following, we describe the construction of the context information $\textbf{Context}$ and provide details on the implementation of the set-to-set function.

\noindent\textbf{How to define the context?} In Eq.~\ref{eq:proj_pred}, the mapping is established between the query instance and the textual information (\ie, classifiers). 
The classifiers represent the typical textual description of the corresponding class, \ie, the common feature. 
Hence, a na\"ive idea is to utilize textual features as the context, \ie,  $\mathbf{W}=[P_t(\w_1), P_t(\w_2), \cdots, P_t(\w_{|\mathcal{Y}_b|})]\in\R^{|\mathcal{Y}_b|\times d}$ (the concatenation of all textual classifiers). However, recent works find an inherent domain gap~\cite{liang2022mind} between the visual and textual embeddings in VLM. 
The gap leads to visual and textual embeddings residing in two separate clusters in the embedding space, which hinders effective pair-wise mapping. 
Correspondingly, we leverage visual prototype features~\cite{snell2017prototypical} as a useful tool for capturing the common characteristics of each class. 
We define the {\em visual prototype} of class $k$ as: 
\begin{equation}	
	\p_k=\frac{1}{N}{\sum_{j=1}^{|\mathcal{{D}}^b|}\mathbb{I}(y_j=k)g_i(\x_j)} \,,
\end{equation}
where $N={\sum_{j=1}^{|\mathcal{{D}}^b|}\mathbb{I}(y_j=k)}$.
They are calculated via forward pass at the beginning of each incremental stage and stay fixed in subsequent tasks. 
Visual prototypes are {\em representative} features of the corresponding class, which can serve as the {\em visual context} to adjust the embeddings. 
Hence, we augment the context with projected visual information, \ie, $[\mathbf{P}, \mathbf{W}]$, where $\mathbf{P}=[P_i(\p_1), P_i(\p_2), \cdots, P_i(\p_{|\mathcal{Y}_b|})]\in \R^{|\mathcal{Y}_b|\times d}$ is the concatenation of all visual prototypes. Combining prototypes from multiple modalities helps the model adapt and fuse information in a cross-modal manner, which goes beyond simple visual-textual matching.

\noindent\textbf{Learning Context Prompts}: In addition to visual prototypes and textual classifiers, we also introduce a set of learnable {\em context prompts} $\{\mathbf{c}^1,\cdots,\mathbf{c}^b\}, \mathbf{c}^i\in \R^{c\times d}$ to be optimized as data evolves. $c$ denotes the length of each prompt. 
Similar to projection layers, we make the context prompts {\em expandable} to catch the new characteristics of new tasks. We initialize a new context prompt while learning a new task and freeze others $\{\bar{\mathbf{c}}^1, \bar{\mathbf{c}}^2,\cdots,{\mathbf{c}^b}\}$. 
The context prompts serve as {\em adaptable} context information, enhancing the co-adaption.  
Hence, the context information is formulated as $\textbf{Context}=[\mathbf{P}, \mathbf{W}, \mathbf{C}]$, where $\mathbf{C}$ is the aggregation of all context prompts. 

\noindent\textbf{Implementing $\mathcal{T}$ with Self-Attention:} In our implementation, we use the self-attention (SA) mechanism~\cite{vaswani2017attention,lin2017structured} as the cross-modal fusion function $\mathcal{T}$. Being permutation invariant, SA is good at outputting adapted embeddings even with long dependencies, which naturally suits the characteristics of the adaptation function.
 Specifically, SA keeps the triplets (query $\mathcal{Q}$, key, $\mathcal{K}$, and value $\mathcal{V}$). The inputs are projected into the same space, \ie, $K=W_{K}^{\top}\left[\,\mathbf{k}_{k}; \, \forall \mathbf{k}_{k} \in \mathcal{K} \,\right] \in \mathbb{R}^{d \times|\mathcal{K}|}$. Similar projections are made for $\mathcal{Q}$ and $\mathcal{V}$. The query $\x_q\in\mathcal{Q}$ is matched against a list of keys $K$ where each key has a value $V$. The output is the sum of all the values weighted by the proximity of the key to the query point:
\begin{align}\label{eq:transformer} \textstyle
	\tilde{P}_i(\z)={P}_i(\z)+\sum_{k} \alpha_{q k} V_{:, k} \,,
\end{align}
where $
\alpha_{q k} \propto \exp \left(\frac{{P}_i(\z)^{\top} W_{Q} \cdot K}{\sqrt{d}}\right)$, $V_{:, k}$ is the $k$-th column of $V$. The adaptation process is the same for other components in $\textbf{Context}$. Specifically, we have $\mathcal{Q}=\mathcal{K}=\mathcal{V}=[P_i(\z), \textbf{Context}]=[P_i(\z),\mathbf{P}, \mathbf{W}, \mathbf{C}]$. The adapted features are then denoted as $[\tilde{P}_i(\z),\tilde{\mathbf{P}}, \tilde{\mathbf{W}}, \tilde{\mathbf{C}}]$ to reflect the context information.

\noindent\textbf{Effect of Cross-Modal Fusion}: 
The illustration of the projection fusion is shown in Figure~\ref{figure:teaser} (right). We utilize the visual and textual information of seen classes as context information to help adjust the {\em instance-specific} embeddings. The fusion model is trained incrementally to adjust embeddings to reflect the context information as data evolves. With the contextualized embeddings, we can conduct the {\em visual matching} and {\em textual matching}:
\begin{align} \label{eq:fuse}
f_\text{VM\&TM}(\x)_{y_i} =&
	\underbrace{
		\frac{\exp \left(\cos \left(\tilde{P}_i(\z), \tilde{P}_i(\p_i)  \right) / \tau\right)}{\sum_{j=1}^{|\mathcal{Y}_b|} \exp \left(\cos \left(\tilde{P}_i(\z), \tilde{P}_i(\p_j) \right) / \tau\right)}
	}_{\text{Visual Matching}}
	+ \\ \notag
	& 
	\underbrace{
		\frac{\exp \left(\cos \left(\tilde{P}_i(\z), \tilde{P}_t(\w_i)  \right) / \tau\right)}{\sum_{j=1}^{|\mathcal{Y}_b|} \exp \left(\cos \left(\tilde{P}_i(\z), \tilde{P}_t(\w_j) \right) / \tau\right)}
	}_{\text{Textual Matching}}
	\,.
\end{align}
In Eq.~\ref{eq:fuse}, the model assigns logits to the query instance by the similarity to the {\em adapted} visual and textual prototypes. 
The incorporation of cross-modal matching enhances the prediction performance.
Note that the context prompt $\mathbf{C}$ only encodes the task-specific information into the self-attention process, \ie, it does not serve as the matching target in Eq.~\ref{eq:fuse}.

\begin{algorithm}[t]
	\caption{Training \name for CIL }
	\label{alg1}
	{\bf Input}: Training dataset: $\D^{b}$; Exemplar set: $\mathcal{E}$; Current model: $f(\cdot)$;\\
	{\bf Output}: Updated model; 
	\begin{algorithmic}[1]
		\State Extract prototypes $\mathbf{p}$ for each new class in $\D^{b}$; \label{line:1}
		\State Freeze current projections and context prompts; \label{line:2}
		\State Initialize new projections  $P_i^b, P_t^b$; {\Comment{\color{cyan}{Expand projections}}}
		\State Initialize new context prompt $\mathbf{c}^b$; \label{line:4}
		\For{$(\x,y) \in \D^b\cup\mathcal{E}$}  {\Comment{\color{cyan}{Incremental learning}}}
		\State Calculate the visual embedding $\z=g_i(\x)$; \label{line:6}
		\State Calculate projected visual/textual embeddings via Eq.~\ref{eq:proj_sum};
		\label{line:9}
		\State Calculate $f_\textbf{PM}(\x)$ via Eq.~\ref{eq:proj_pred};{\Comment{\color{cyan}{Projected matching}}} \label{line:10}
		\State Cross-modal fusion via Eq.~\ref{eq:transformer}; {\Comment{\color{cyan}{Cross-modal fusion}}} \label{line:11}
		\State Calculate visual and textual matching via Eq.~\ref{eq:fuse}; \label{line:13}
		\State Calculate the loss via Eq.~\ref{eq:loss}; update the model; \label{line:14}
		\EndFor 
		\Return the updated model;
	\end{algorithmic}
\end{algorithm}

\subsection{Summary of \name}
In \mame, we first enable learning new concepts via projected mapping. Then, to accommodate new concepts without interference from previous ones, we initialize new projections for each new task and freeze the former ones. Besides, we utilize self-attention to adjust the embeddings of the query instance and the context information to promote cross-modal fusion. 
Figure~\ref{figure:teaser} illustrates three matching targets, \ie, projected matching (Eq.~\ref{eq:proj_pred}), visual/textual matching (Eq.~\ref{eq:fuse}).
We denote these models as $f_\text{PM}(\x), f_\text{VM}(\x), f_\text{TM}(\x)$, respectively.
 During training, we optimize the cross-entropy loss:
\begin{equation} \label{eq:loss}  
	\min_{\{P_i^b,P_t^b,\mathcal{T},\mathbf{c}^b \}} \ell(f_\text{PM}(\x),y) + \ell(f_\text{VM}(\x),y) +\ell(f_\text{TM}(\x),y) \,,
\end{equation}
where $(\x,y) \in \D^b \cup \mathcal{E}$.
In Eq.~\ref{eq:loss}, all pre-trained weights are frozen, and we only optimize these {\em additional} parameters. For inference, we aggregate the three logits, \ie, $f(\x)=f_\text{PM}(\x)+f_\text{VM}(\x)+f_\text{TM}(\x)$.

\begin{table*}[!htb]
	\caption{Average and last performance comparison of different methods.  
		The first and second columns represent the methods with and without exemplars.
		The performance of L2P, DualPrompt, CODA-Prompt, PLOT, and DAP are reproduced with the source code with exemplars using CLIP's visual branch. The best performance is shown in bold.  All methods are initialized with the same pre-trained CLIP for a fair comparison. All exemplar-related methods utilize the same number of exemplars for a fair comparison.
	}\label{tab:supp_benchmark}
	\centering
	\resizebox{1.0\textwidth}{!}{%
		\begin{tabular}{@{}lccccccccccccccc}
			\toprule
			\multicolumn{1}{c}{\multirow{3}{*}{Method}}
			&
			\multicolumn{1}{c}{\multirow{3}{*}{Exemplar}}
			& 
			\multicolumn{4}{c}{Aircraft }   & 
			\multicolumn{4}{c}{CIFAR100 }	&	\multicolumn{4}{c}{Cars }   
			\\ 
			& & 
			\multicolumn{2}{c}{B0 Inc10}   & 
			\multicolumn{2}{c}{B50 Inc10}	&		\multicolumn{2}{c}{B0 Inc10}   & 
			\multicolumn{2}{c}{B50 Inc10}	& 
			\multicolumn{2}{c}{B0 Inc10}   & 
			\multicolumn{2}{c}{B50 Inc10}	& 
			\\  
			& & 
			{$\bar{\mathcal{A}}$} & ${\mathcal{A}_B}$  
			& {$\bar{\mathcal{A}}$} & ${\mathcal{A}_B}$
			& {$\bar{\mathcal{A}}$} & ${\mathcal{A}_B}$ 
			&  {$\bar{\mathcal{A}}$} & ${\mathcal{A}_B}$  
			& {$\bar{\mathcal{A}}$} & ${\mathcal{A}_B}$
			& {$\bar{\mathcal{A}}$} & ${\mathcal{A}_B}$ 
			\\
			\midrule
			Oracle & & \multicolumn{4}{c}{61.12}& \multicolumn{4}{c}{82.35} & \multicolumn{4}{c}{90.42} \\
			\midrule
			Finetune & \ding{55} &  3.16 & 0.96 & 1.72 & 1.05 & 7.84 & 4.44& 5.30 & 2.46& 3.14 & 1.10 & 1.54 & 1.13\\
			Finetune LiT~\cite{zhai2022lit}& \ding{55} & 27.74 & 14.28  & 25.10 & 13.77 & 44.66 & 14.69& 27.69 & 7.67& 84.12 & 72.37& 83.08 & 78.23\\
			Finetune CoOp~\cite{zhou2022learning}& \ding{55} & 14.54 & 7.14 & 13.05 & 7.77 & 47.00 & 24.24 & 41.23 & 24.12& 36.46 & 21.65& 37.40 & 20.87\\
			SimpleCIL~\cite{zhou2023revisiting}& \ding{55} &59.24 & 48.09 & 53.05 & 48.09 & 84.15 & 76.63& 80.20 & 76.63& 92.04 & 86.85 & 88.96 & 86.85\\
			ZS-CLIP~\cite{radford2021learning}& \ding{55} &26.66 & 17.22 & 21.70 & 17.22& 81.81 & 71.38& 76.49 & 71.38& 82.60 & 76.37& 78.32 & 76.37\\
			\midrule
			CoOp~\cite{zhou2022learning}& \ding{51} &44.26 & 39.87 & 41.81 & 39.18& 83.37 & 73.36& 78.34 & 73.04& 89.73 & 84.91& 87.98 & 86.60\\
			iCaRL~\cite{rebuffi2017icarl}& \ding{51} &  53.60 & 43.98  & 50.40 & 45.33& 79.91 & 63.94& 71.94 & 63.00& 89.38 & 84.95& 86.71 & 84.19\\
			MEMO~\cite{zhou2022model}& \ding{51} & 42.24 & 25.41 & 38.16 & 27.75& 84.67 & 74.98 & 80.75 & 75.34 & 88.23 & 81.31& 84.90 & 81.83\\
			L2P~\cite{wang2022learning}  & \ding{51}  &  55.06 & 44.88 & 47.78 & 43.37 & 76.42 & 66.21 & 72.67 & 67.88& 83.81 & 72.44& 79.76 & 73.47\\
			DualPrompt~\cite{wang2022dualprompt} & \ding{51}    &55.95 & 46.53 & 50.93 & 46.50& 79.07 & 70.06&   74.81 & 70.75 & 85.30 & 74.35 & 81.32 & 75.85\\
			CODA-Prompt~\cite{smith2023coda} & \ding{51}    &63.13 & 52.27 & 62.05 & 54.70& 82.91 & 74.23&  81.33 & 75.92 & 92.11 & 89.20 & 89.45 & 87.84
			\\
			DAP~\cite{jung2023generating} & \ding{51}    &29.02 & 10.92 & 41.45 & 28.56& 68.10 & 40.80&  76.57 & 59.92 & 84.87 & 81.31 & 84.63 & 83.15\\
			PLOT~\cite{chen2023plot} & \ding{51}    &50.47 & 43.65 & 46.82 & 43.58& 72.62 & 56.81&  74.35 & 67.90 & 86.54 & 83.45 & 82.43 & 74.26\\	
			\rowcolor{LightCyan}\name  & \ding{51}  & \bf 64.61 & \bf 55.81 & \bf 63.59 & \bf 58.81 & \bf86.70 &\bf 79.05 & \bf82.92 & \bf78.87& \bf93.26 &\bf 89.84 & \bf90.53 & \bf89.54\\
		\end{tabular}
	}	
	
	\resizebox{1.0\textwidth}{!}{%
		\begin{tabular}{@{}lccccccccccccccc}
			\toprule
			\multicolumn{1}{c}{\multirow{3}{*}{Method}}
			&
			\multicolumn{1}{c}{\multirow{3}{*}{Exemplar}}
			& 
			\multicolumn{4}{c}{ImageNet-R }   & 
			\multicolumn{4}{c}{CUB }	&	\multicolumn{4}{c}{UCF }   
			\\ 
			& & 
			\multicolumn{2}{c}{B0 Inc20}   & 
			\multicolumn{2}{c}{B100 Inc20}	&	\multicolumn{2}{c}{B0 Inc20}   & 
			\multicolumn{2}{c}{B100 Inc20}	& 
			\multicolumn{2}{c}{B0 Inc10}   & 
			\multicolumn{2}{c}{B50 Inc10}	& 
			\\  
			& & 
			{$\bar{\mathcal{A}}$} & ${\mathcal{A}_B}$  
			& {$\bar{\mathcal{A}}$} & ${\mathcal{A}_B}$
			& {$\bar{\mathcal{A}}$} & ${\mathcal{A}_B}$ 
			&  {$\bar{\mathcal{A}}$} & ${\mathcal{A}_B}$  
			& {$\bar{\mathcal{A}}$} & ${\mathcal{A}_B}$
			& {$\bar{\mathcal{A}}$} & ${\mathcal{A}_B}$ 
			\\
			\midrule
			Oracle & & \multicolumn{4}{c}{80.28}& \multicolumn{4}{c}{79.47} & \multicolumn{4}{c}{93.63}  \\
			\midrule
			Finetune & \ding{55} & 1.37 & 0.43& 1.01 & 0.88& 2.06 & 0.64& 0.56 & 0.47& 4.51 & 1.59& 1.21 & 0.80\\
			Finetune LiT~\cite{zhai2022lit}& \ding{55} & 64.88 & 30.42& 57.75 & 29.77& 58.15 & 35.28& 51.95 & 35.96& 79.25 & 64.84& 81.79 & 65.40\\
			Finetune CoOp~\cite{zhou2022learning}& \ding{55} &60.73 & 37.52& 54.20 & 39.77& 27.61 & 8.57& 24.03 & 10.14& 47.85 & 33.46& 42.02 & 24.74\\
			SimpleCIL~\cite{zhou2023revisiting}& \ding{55} & 81.06 & 74.48& 76.84 & 74.48& 83.81 & 77.52& 79.75 & 77.52& 90.44 & 85.68& 88.12 & 85.68\\
			ZS-CLIP~\cite{radford2021learning}& \ding{55} &83.37 & 77.17& 79.57 & 77.17 & 74.38 & 63.06& 67.96 & 63.06& 75.50 & 67.64& 71.44 & 67.64\\
			\midrule
			CoOp~\cite{zhou2022learning}& \ding{51} &82.40 & 76.20& 79.76 & 77.13& 77.34 & 68.70& 74.09 & 67.47& 90.13 & 86.24& 88.36 & 85.71\\
			iCaRL~\cite{rebuffi2017icarl}& \ding{51} & 72.22 & 54.38& 68.67 & 60.15& 82.04 & 74.74& 78.57 & 75.07 & 89.47 & 84.34& 88.51 & 84.11\\
			MEMO~\cite{zhou2022model}& \ding{51} & 80.00 & 74.07  & 76.72 & 73.95& 77.32 & 65.69 & 72.88 & 66.41& 84.02 & 74.08& 82.58 & 75.48\\
			L2P~\cite{wang2022learning}  & \ding{51}  &   75.73 & 67.22& 74.15 & 71.20& 79.23 & 68.54& 75.85 & 71.12& 88.71 & 83.93& 86.51 & 83.22\\
			DualPrompt~\cite{wang2022dualprompt} & \ding{51}    &78.47 & 70.82&  72.98 & 69.18 & 83.21 & 74.94&  78.06 & 74.27& 89.48 & 85.41& 86.96 & 84.65\\
			CODA-Prompt~\cite{smith2023coda} & \ding{51}     &80.91 & 74.20 & 79.32 & 74.73& 82.91 & 73.20&  79.81 & 74.73 & 92.51 & 89.74 & 92.73 & 90.28\\
			DAP~\cite{jung2023generating} & \ding{51}    &78.00 & 72.73 & 77.23 & 74.37& 76.48 & 73.07&  75.39 & 74.09 & 87.63 & 81.81 & 87.64 & 85.68\\
			PLOT~\cite{chen2023plot} & \ding{51}   &72.67 & 66.52 & 70.45 & 68.24& 80.46 & 71.34&  78.35 & 72.03 & 83.54 & 73.78 & 87.09 & 82.91\\
			\rowcolor{LightCyan}\name  & \ding{51}  &  \bf85.34 &\bf 80.10& \bf82.32 & \bf80.30 & \bf84.93 &\bf 79.43&\bf 81.67 &\bf 79.18& \bf94.34 &\bf 90.60&\bf 93.56 & \bf91.32\\
		\end{tabular}
	}
	
	\resizebox{1.0\textwidth}{!}{%
		\begin{tabular}{@{}lccccccccccccccc}
			\toprule
			\multicolumn{1}{c}{\multirow{3}{*}{Method}}
			&
			\multicolumn{1}{c}{\multirow{3}{*}{Exemplar}}
			& 
			\multicolumn{4}{c}{SUN }   & 
			\multicolumn{4}{c}{Food }	&	\multicolumn{4}{c}{ObjectNet }   
			\\ 
			& & 
			\multicolumn{2}{c}{B0 Inc30}   & 
			\multicolumn{2}{c}{B150 Inc30}	&		\multicolumn{2}{c}{B0 Inc10}   & 
			\multicolumn{2}{c}{B50 Inc10}	& 
			\multicolumn{2}{c}{B0 Inc20}   & 
			\multicolumn{2}{c}{B100 Inc20}	& 
			\\  
			& & 
			{$\bar{\mathcal{A}}$} & ${\mathcal{A}_B}$  
			& {$\bar{\mathcal{A}}$} & ${\mathcal{A}_B}$
			& {$\bar{\mathcal{A}}$} & ${\mathcal{A}_B}$ 
			&  {$\bar{\mathcal{A}}$} & ${\mathcal{A}_B}$  
			& {$\bar{\mathcal{A}}$} & ${\mathcal{A}_B}$
			& {$\bar{\mathcal{A}}$} & ${\mathcal{A}_B}$ 
			\\
			\midrule
			Oracle& & \multicolumn{4}{c}{81.74}& \multicolumn{4}{c}{85.83} & \multicolumn{4}{c}{45.46}  \\
			\midrule
			Finetune & \ding{55} &4.51 & 1.59& 0.78 & 0.72& 3.49 & 1.71& 2.14 & 1.52& 1.34 & 0.47 & 0.69 & 0.54\\
			Finetune LiT~\cite{zhai2022lit}& \ding{55} & 79.25 & 64.84& 38.23 & 20.00& 40.62 & 12.96& 29.74 & 12.05& 43.27 & 17.46& 32.85 & 17.17\\
			Finetune CoOp~\cite{zhou2022learning}& \ding{55} &45.93 & 23.11 & 39.33 & 24.89& 36.01 & 14.18& 33.13 & 18.67& 21.24 & 6.29& 16.21 & 6.82\\
			SimpleCIL~\cite{zhou2023revisiting}& \ding{55} & 82.13 & 75.58& 78.62 & 75.58& 87.89 & 81.65& 84.73 & 81.65& 52.06 & 40.13& 45.11 & 40.13\\
			ZS-CLIP~\cite{radford2021learning}& \ding{55} &79.42 & 72.11& 74.95 & 72.11& 87.86 & 81.92& 84.75 & 81.92& 38.43 & 26.43 & 31.12 & 26.43\\
			\midrule
			CoOp~\cite{zhou2022learning}& \ding{51} &80.46 & 73.44& 77.68 & 73.06& 85.38 & 76.15& 81.74 & 76.35& 46.16 & 33.81& 40.40 & 34.47\\
			iCaRL~\cite{rebuffi2017icarl}& \ding{51} &78.56 & 67.30 & 74.74 & 69.07 & 84.12 & 71.68& 78.86 & 70.64& 45.28 & 26.97 & 37.22 & 26.15\\
			MEMO~\cite{zhou2022model}& \ding{51}&81.48 & 73.45 & 78.00 & 73.87 & 89.18 & 82.85& 86.50 & 83.08& 46.98 & 33.37& 41.62 & 34.67\\
			L2P~\cite{wang2022learning}  & \ding{51}  &   79.83 & 72.14 & 76.16 & 72.32& 84.48 & 75.22& 85.04 & 80.56& 46.18 & 34.00& 43.90 & 39.57\\
			DualPrompt~\cite{wang2022dualprompt} & \ding{51}    &80.14 & 73.06& 77.25 & 73.82& 87.12 & 81.27 & 85.37 & 82.36  & 53.13 & 40.59& 45.84 & 40.37\\
			CODA-Prompt~\cite{smith2023coda} & \ding{51}   &80.91 & 73.23 & 78.38 & 69.45& 80.82 & 72.81&  79.35 & 73.46 & 53.38 & 41.12 & 47.86 & 42.35\\
			DAP~\cite{jung2023generating} & \ding{51}    &79.81 & 73.89 & 79.61 & 74.78& 78.49 & 71.37&  81.68 & 78.38 & 40.10 & 37.09 & 42.47 & 32.95\\
			PLOT~\cite{chen2023plot} & \ding{51}   &74.91 & 62.21 & 75.34 & 68.59& 82.37 & 74.34&  78.39 & 72.49 & 45.34 & 34.85 & 41.85 & 33.38\\
			\rowcolor{LightCyan}\name  & \ding{51}  & \bf83.57 & \bf77.28& \bf80.70 & \bf77.49& \bf90.04 & \bf84.73& \bf87.52 &\bf 84.74&\bf 55.28 &\bf 44.36& \bf49.64 &\bf 43.65 \\
			\bottomrule
		\end{tabular}
	}
\end{table*}

\begin{figure*}[t]
	\begin{center}
		\subfigure[ Aircraft Base0 Inc10]
		{\includegraphics[width=.65\columnwidth]{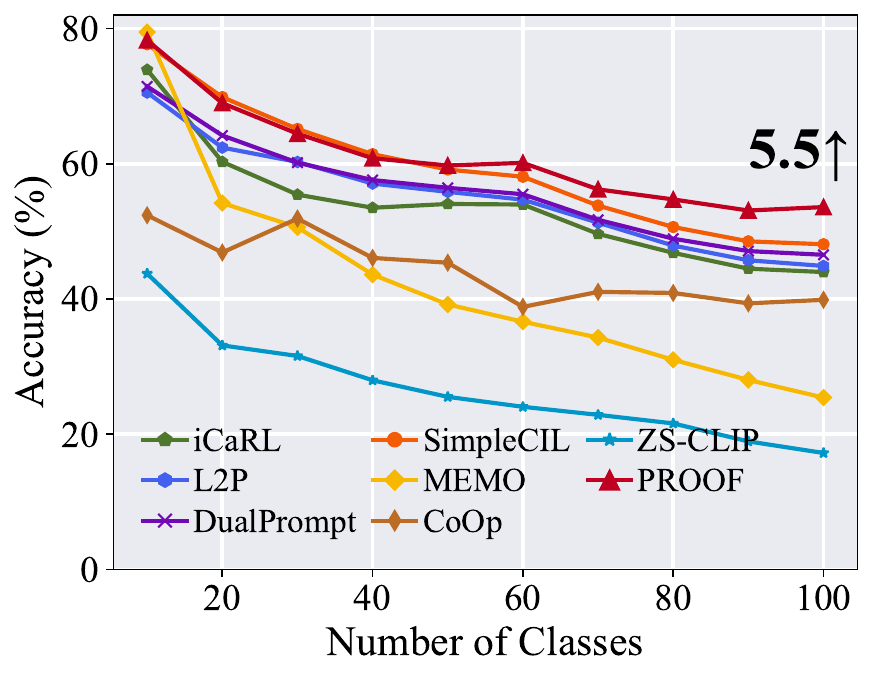}}
		\subfigure[CIFAR100 Base0 Inc10]
		{\includegraphics[width=.65\columnwidth]{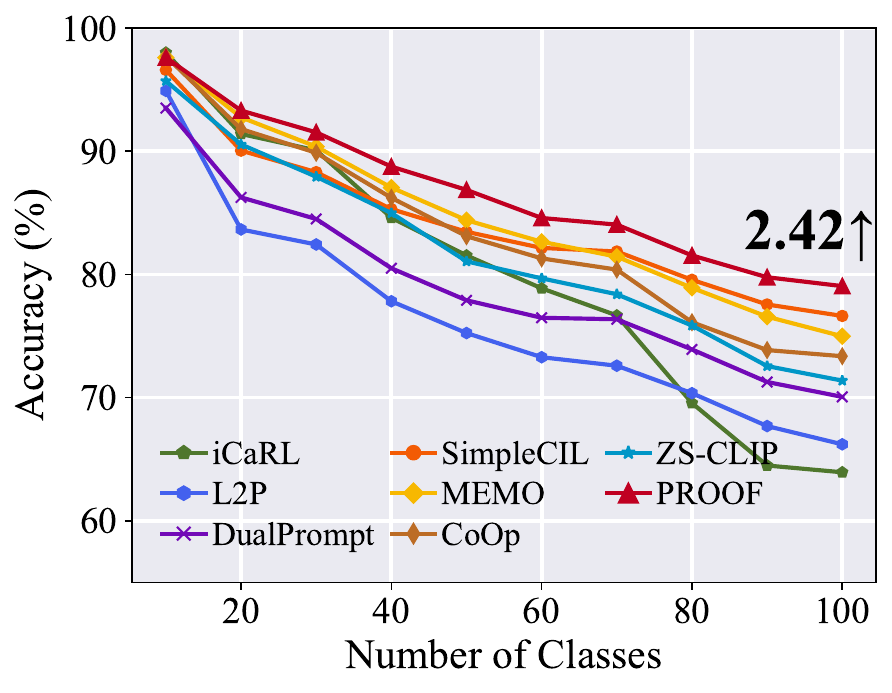}}
		\subfigure[Cars Base0 Inc10]
		{\includegraphics[width=.65\columnwidth]{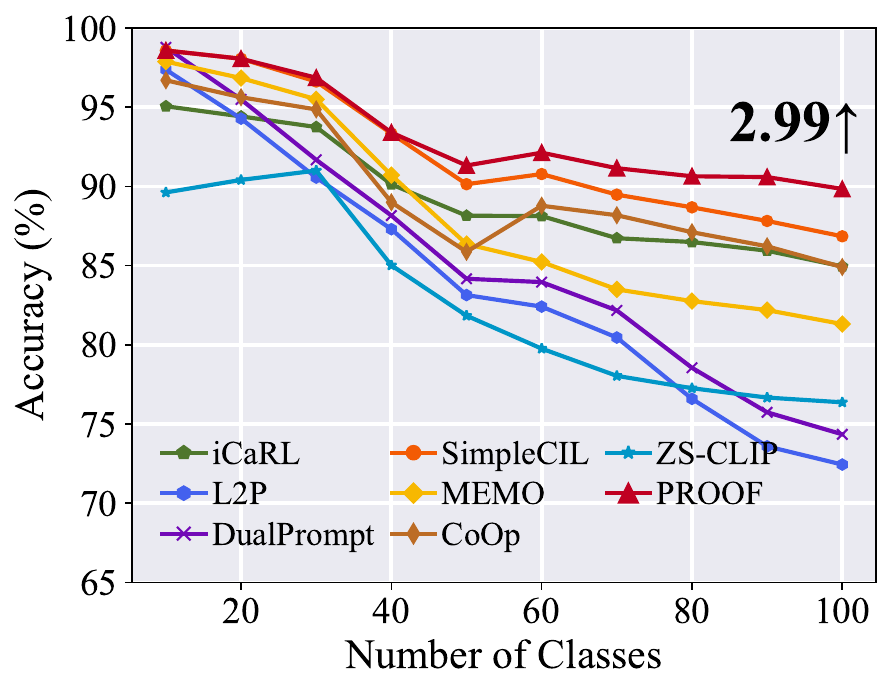}}\\
		\subfigure[ImageNet-R Base0 Inc20]
		{\includegraphics[width=.65\columnwidth]{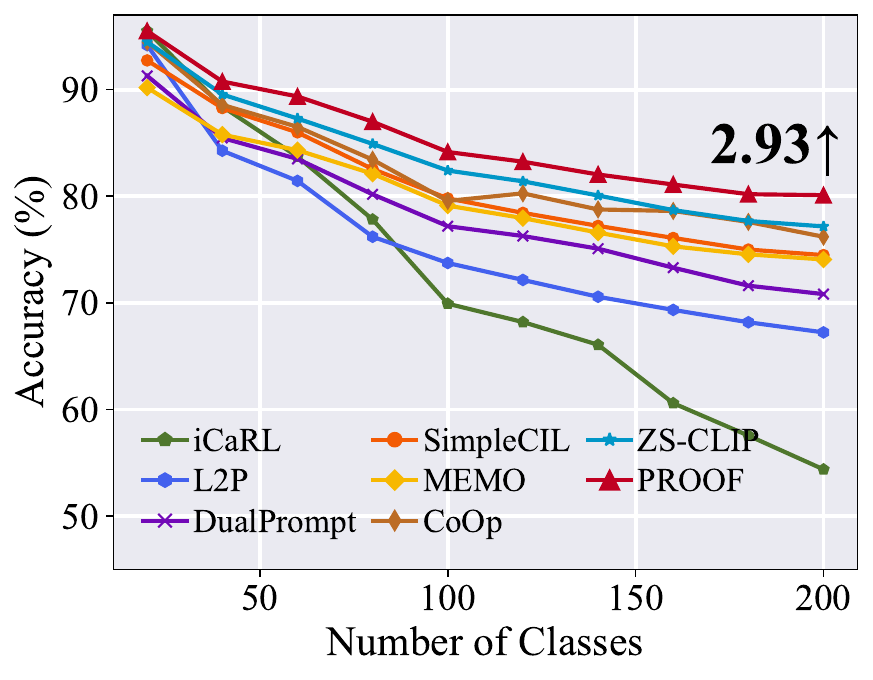}}
		\subfigure[CUB Base0 Inc20]
		{\includegraphics[width=.65\columnwidth]{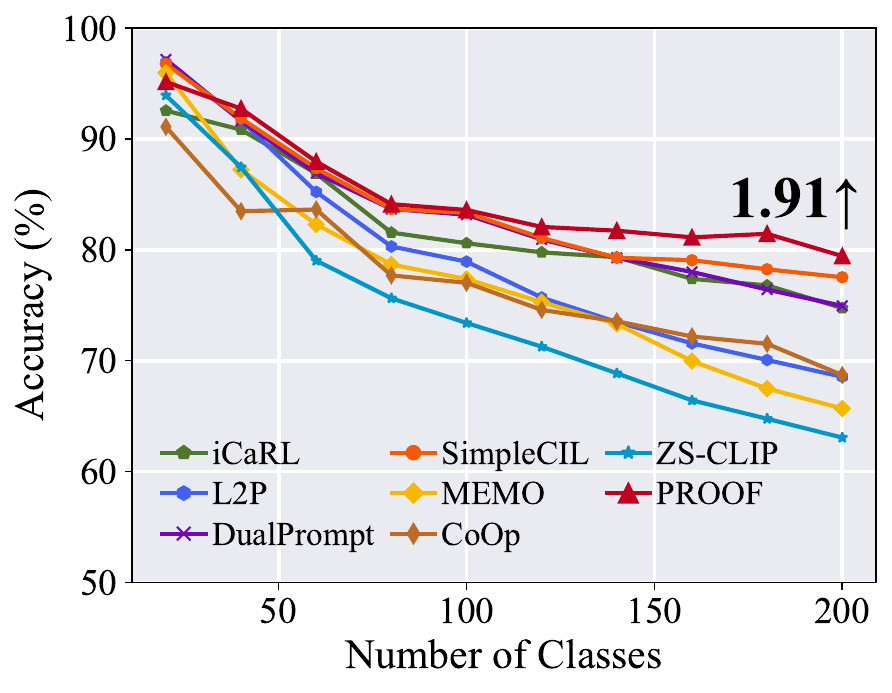}}
		\subfigure[UCF Base0 Inc10]
		{\includegraphics[width=.65\columnwidth]{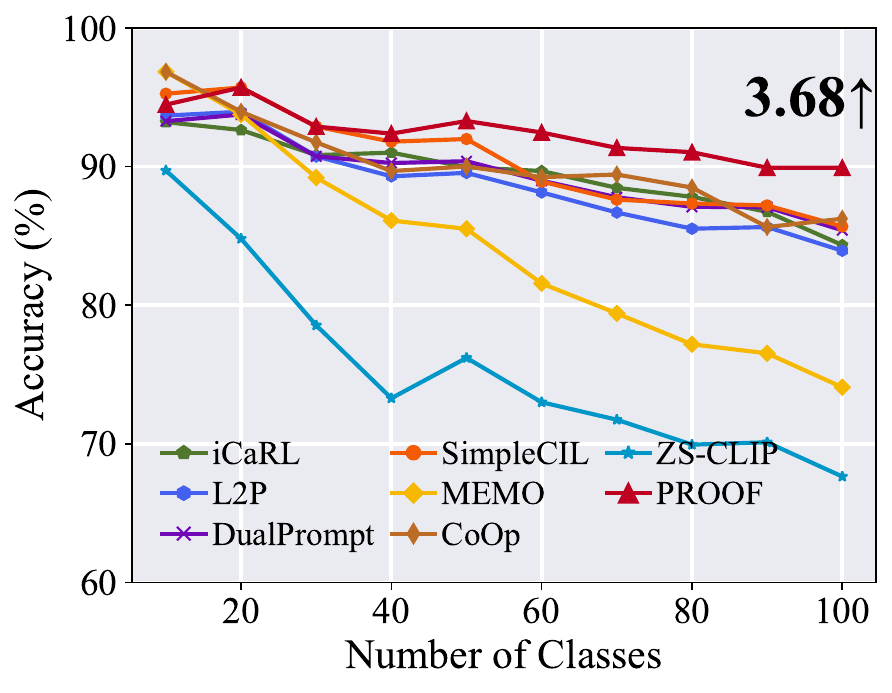}}\\
		\subfigure[SUN Base0 Inc30]
		{\includegraphics[width=.65\columnwidth]{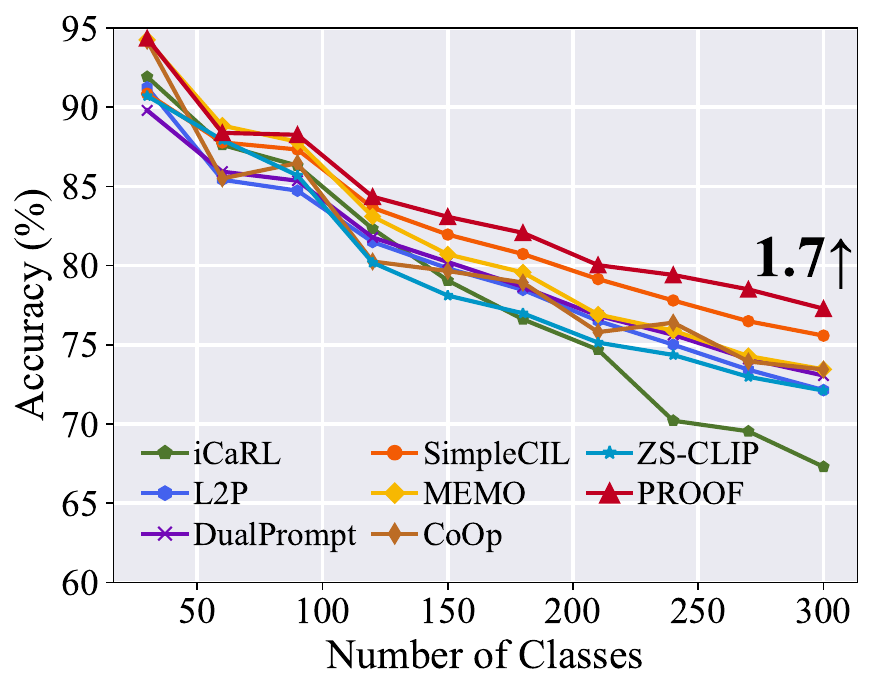}}
		\subfigure[Food Base0 Inc10]
		{\includegraphics[width=.65\columnwidth]{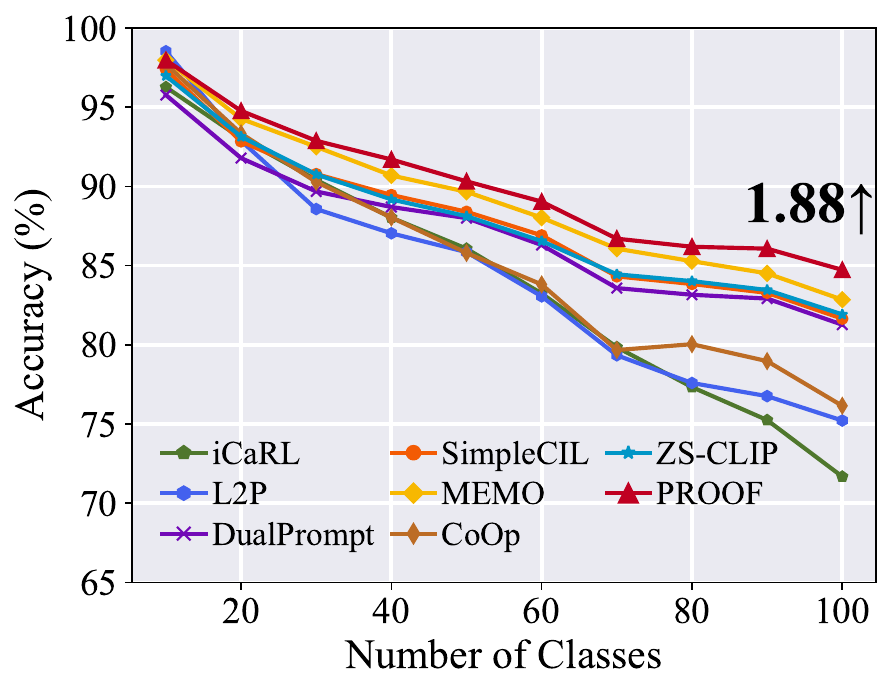}}
		\subfigure[ObjectNet Base0 Inc20]
		{\includegraphics[width=.65\columnwidth]{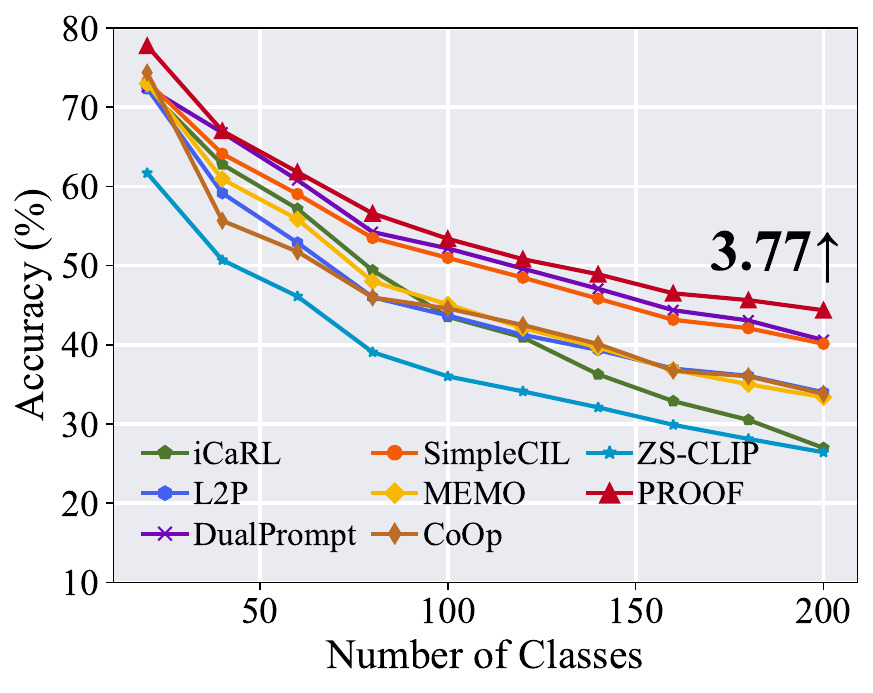}}\\
	\end{center}
	\caption{
		Incremental performance of different methods. We report the performance gap after the last incremental stage of \name and the runner-up method at the end of the line.    All methods are based on the same backbone/weight.
	}
	
	\label{figure:supp_benchmark}
	
\end{figure*}

\begin{figure*}[t]
	\begin{center}
		\subfigure[ Aircraft Base50 Inc10]
		{\includegraphics[width=.65\columnwidth]{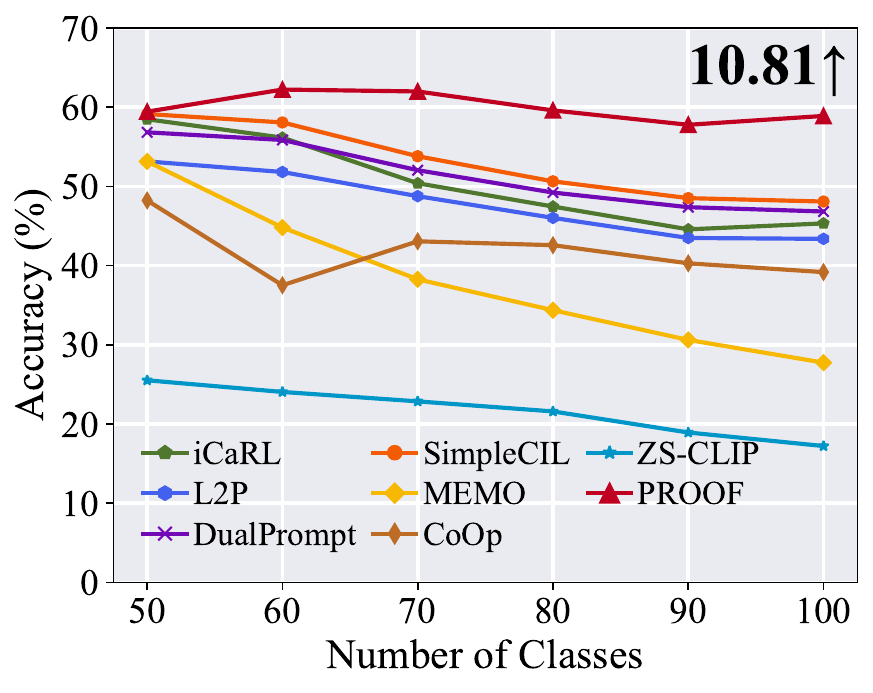}}
		\subfigure[CIFAR100 Base50 Inc10]
		{\includegraphics[width=.65\columnwidth]{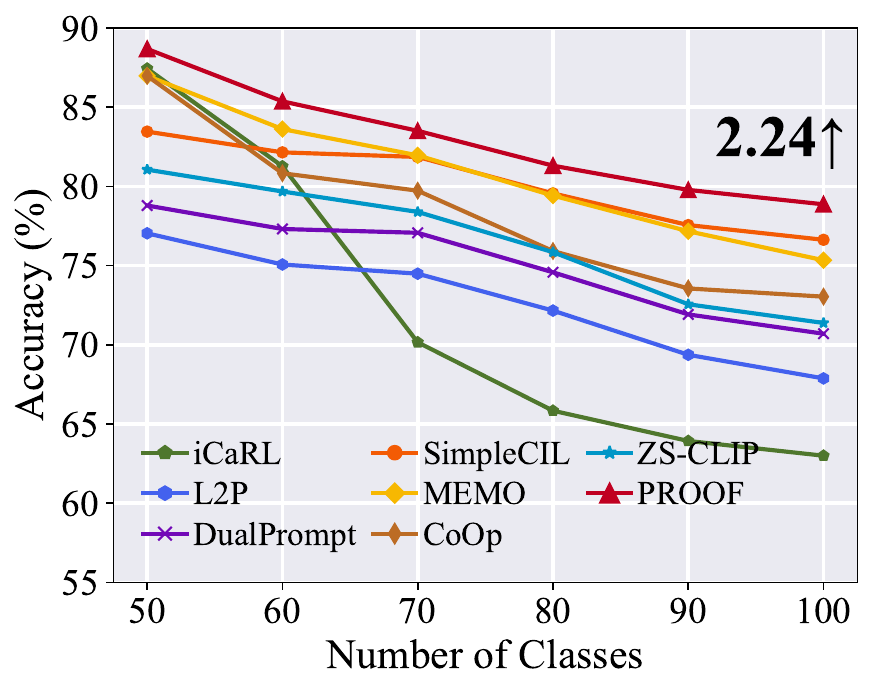}}
		\subfigure[Cars Base50 Inc10]
		{\includegraphics[width=.65\columnwidth]{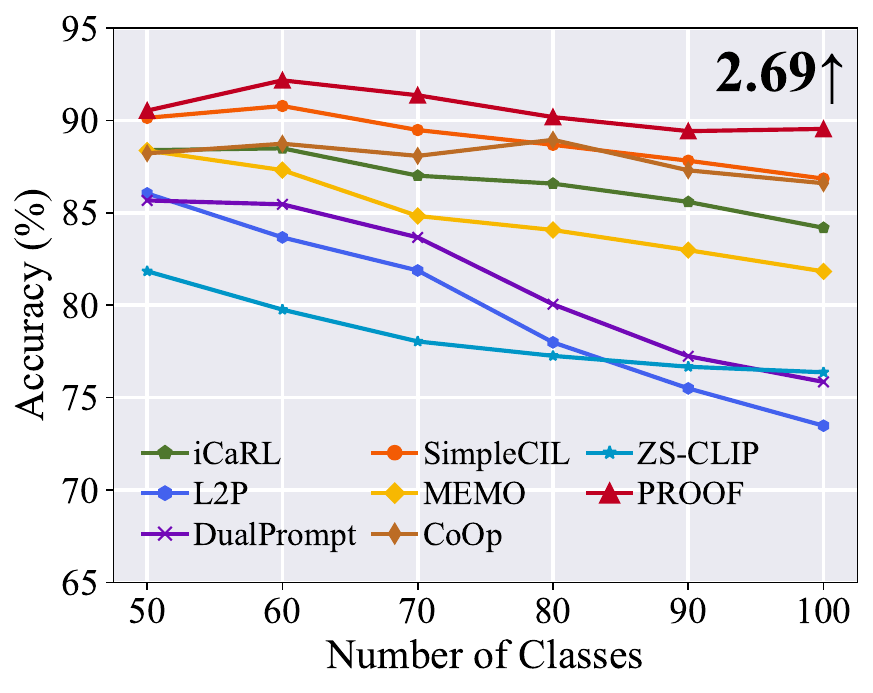}}\\
		\subfigure[ImageNet-R Base100 Inc20]
		{\includegraphics[width=.65\columnwidth]{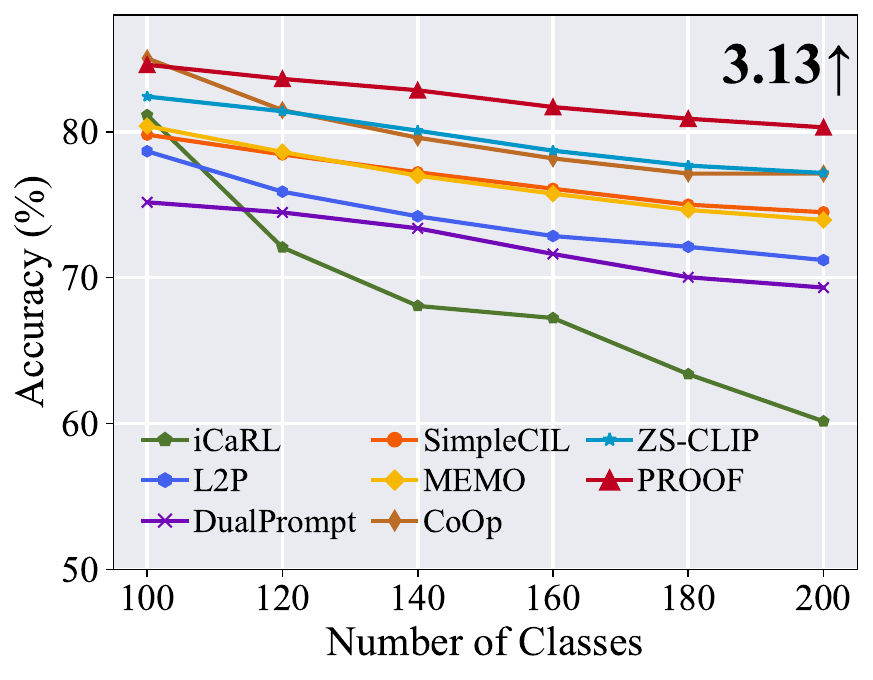}}
		\subfigure[CUB Base100 Inc20]
		{\includegraphics[width=.65\columnwidth]{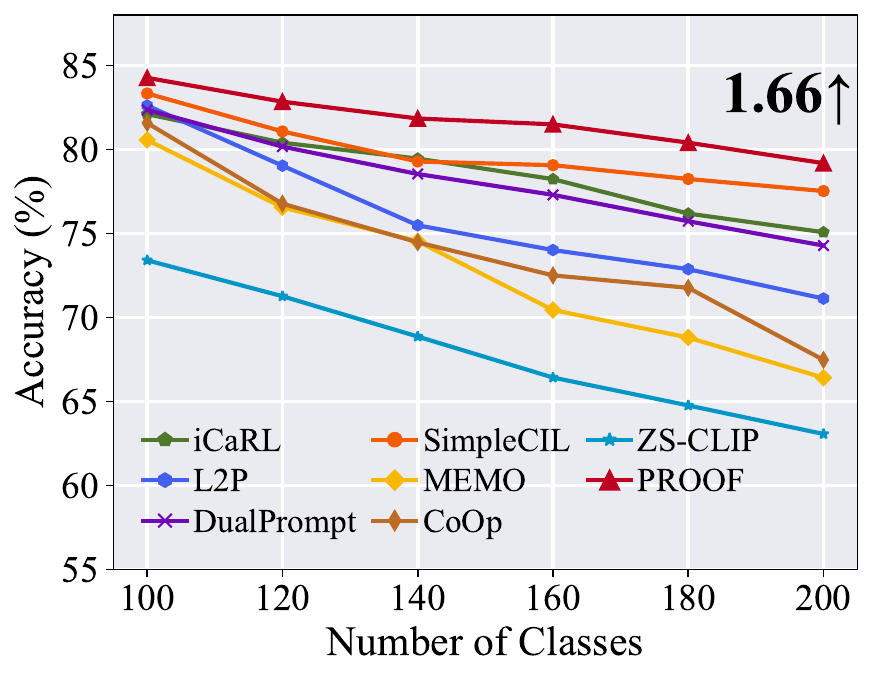}}
		\subfigure[UCF Base50 Inc10]
		{\includegraphics[width=.65\columnwidth]{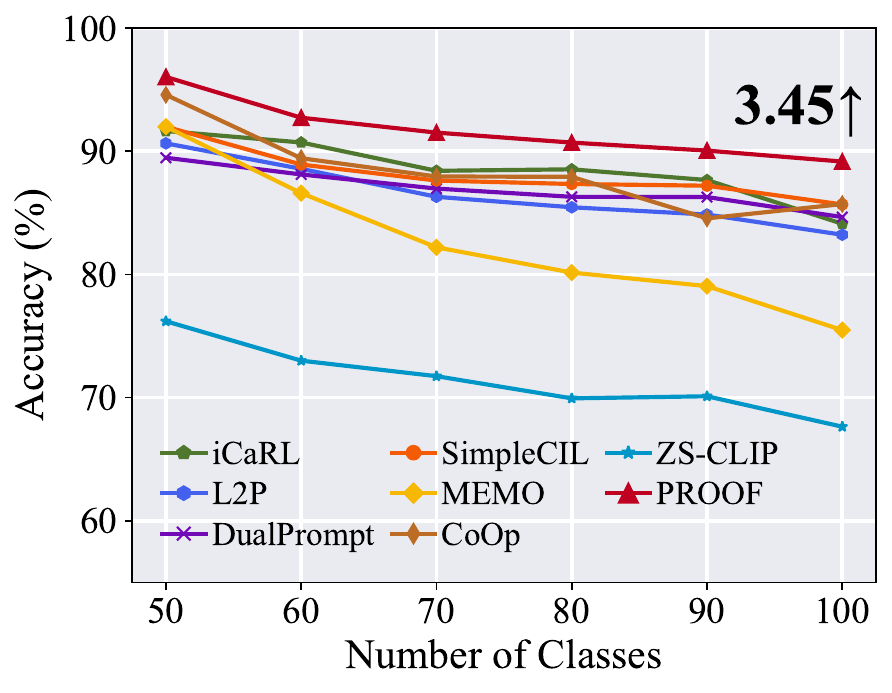}}\\
		\subfigure[SUN Base150 Inc30]
		{\includegraphics[width=.65\columnwidth]{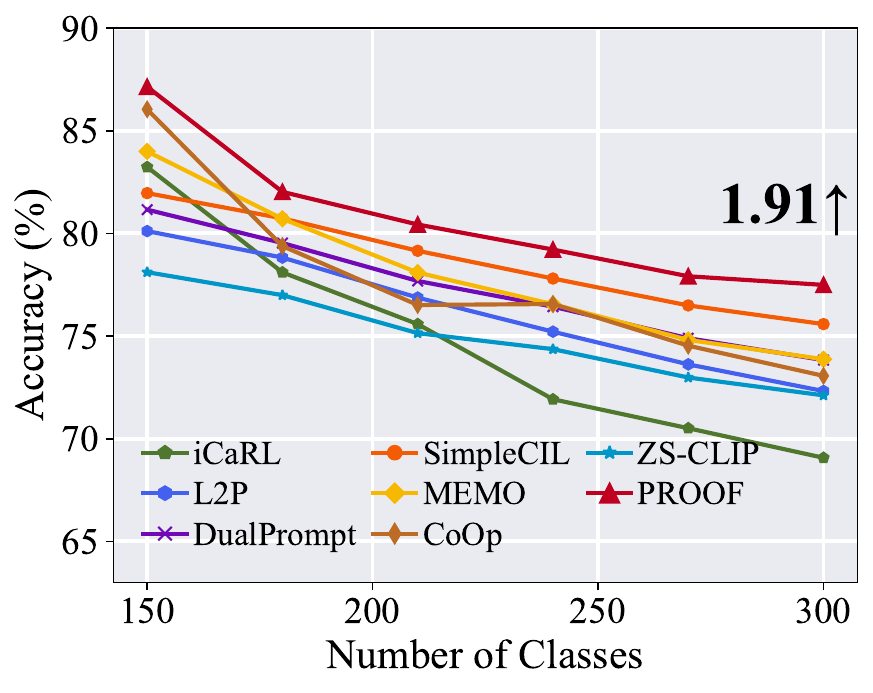}}
		\subfigure[Food Base50 Inc10]
		{\includegraphics[width=.65\columnwidth]{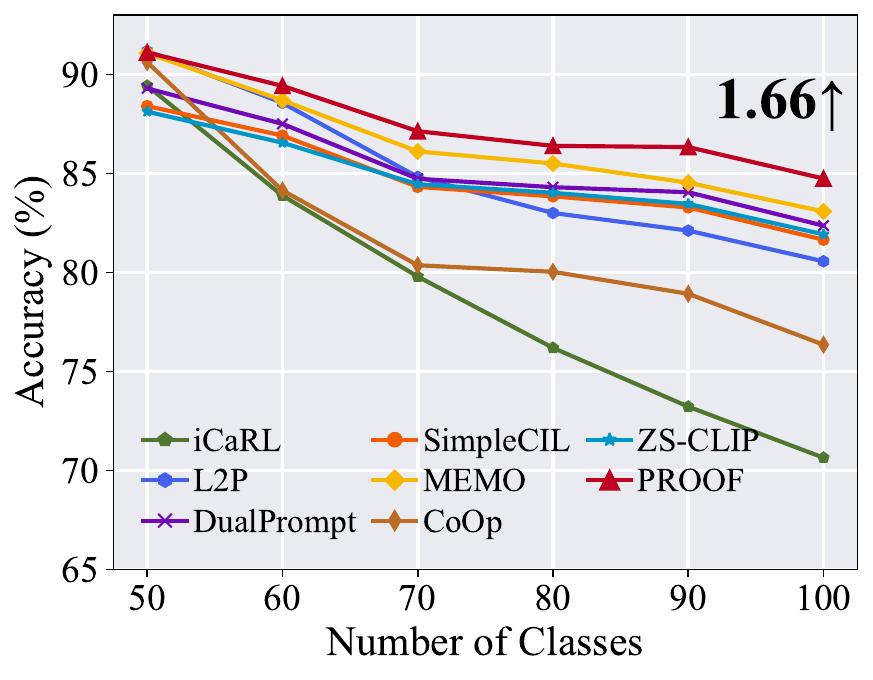}}
		\subfigure[ObjectNet Base100 Inc20]
		{\includegraphics[width=.65\columnwidth]{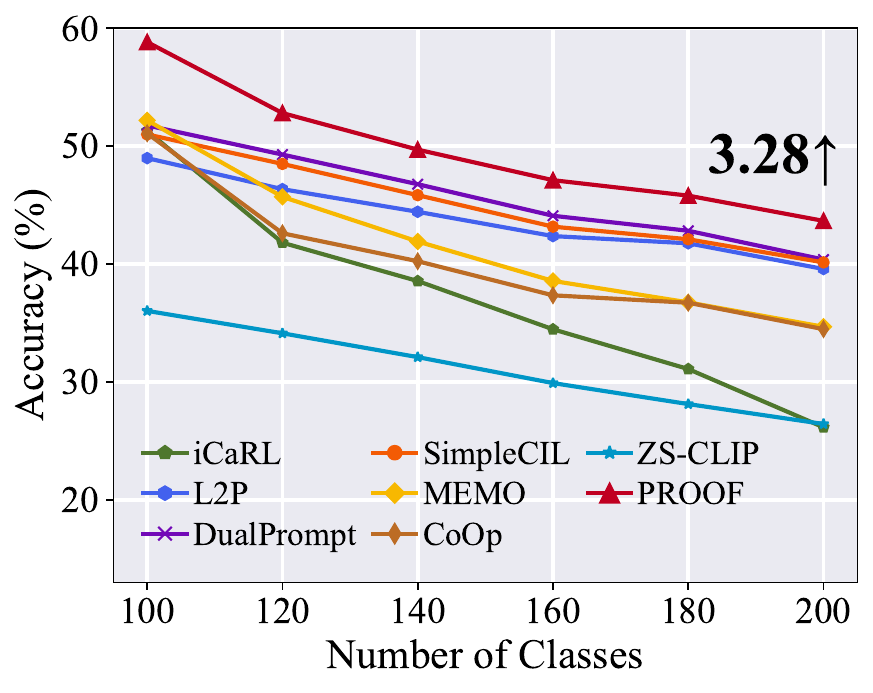}}
	\end{center}
	\caption{
		Incremental performance of different methods with large base classes. We report the performance gap after the last incremental stage of \name and the runner-up method at the end of the line.   All methods are based on the same backbone/weight.
	}
	\label{figure:benchmark_large_base}
\end{figure*}

\noindent\textbf{Pseudo Code:} We give the pseudo-code of \name to illustrate the training process in Algorithm~\ref{alg1}.
In each incremental stage, we are provided with the training dataset $\D^{b}$ and the exemplar set $\mathcal{E}$ to update the current model $f(\cdot)$. Before training, we initially extract visual prototypes for the new classes. These prototypes are calculated using the frozen visual embedding $g_i(\cdot)$, ensuring their stability throughout model updates. Subsequently, we freeze the former projections and context prompts while initializing new projections and context prompts specifically for the new incremental task (Line~\ref{line:2} to Line~\ref{line:4}). These steps represent the model expansion process, which is followed by the subsequent learning process.

During the learning process, we concatenate the training instances from the current dataset and the exemplar set, initiating a for-loop. For each instance-label pair, we calculate the projected visual and textual embeddings (Line~\ref{line:6} to Line~\ref{line:9}). Subsequently, we compute the projected matching loss (Line~\ref{line:10}) to encode task-specific information into the current projection layers. Based on the projected features, we derive context information and perform cross-modal fusion (Line~\ref{line:11} to Line~\ref{line:13}). Consequently, we obtain three logits for model updating and utilize the cross-entropy loss to update these modules (Line~\ref{line:14}). The updated model is then returned as the output of the training process.


\section{Experiment}
In this section, we compare \name to state-of-the-art methods on benchmark datasets to investigate the capability of overcoming forgetting. Besides, we conduct ablations to analyze the effect of each component in the model. 
We also extend \name to other VLMs and continual learning scenarios, experiment with a non-overlapping dataset, and address the zero-shot performance degradation phenomena.

\subsection{Experimental Setup}\label{sec:setup}
\noindent {\bf Dataset}: Following the benchmark CIL settings~\cite{rebuffi2017icarl,wang2022learning,wang2022dualprompt,yu2020semantic,zhou2023revisiting}, we evaluate the performance on \bfname{CIFAR100}~\cite{krizhevsky2009learning}, \bfname{CUB200}~\cite{WahCUB2002011}, \bfname{ObjectNet}~\cite{barbu2019objectnet}, and \bfname{ImageNet-R}~\cite{hendrycks2021many}. We also follow the benchmark in VLM tuning~\cite{zhou2022learning}, and formulate \bfname{FGVCAircraft}~\cite{maji2013fine}, \bfname{StanfordCars}~\cite{krause20133d}, \bfname{Food101}~\cite{bossard2014food}, \bfname{SUN397}~\cite{xiao2010sun} and \bfname{UCF101}~\cite{soomro2012ucf101} into CIL setting. 
Specifically, we sample (a subset of) 100 classes from CIFAR100, Aircraft, Cars, Food, UCF, 200 classes from CUB200, ObjectNet, ImageNet-R, and 300 classes from SUN to ease the data split. 
Following~\cite{rebuffi2017icarl}, the training class order is shuffled with random seed 1993. The dataset splits are denoted as \bfname{Base-$x$, Inc-$y$}, where $x$ represents the number of classes in the first stage, and $y$ represents the number of new classes in each subsequent task. 
$x=0$ means each task contains $y$ classes.

\noindent {\bf Comparison methods:} We first compare to SOTA CIL methods iCaRL~\cite{rebuffi2017icarl},  MEMO~\cite{zhou2022model},  SimpleCIL~\cite{zhou2023revisiting}, L2P~\cite{wang2022learning}, DualPrompt~\cite{wang2022dualprompt}. Denote the baseline of sequential finetuning as Finetune; we combine it with different tuning techniques, \eg, LiT~\cite{zhai2022lit}, PLOT~\cite{chen2023plot}, and CoOp~\cite{zhou2022learning}. We also report the zero-shot performance of CLIP as ZS-CLIP by matching the query instance to the template (Eq.~\ref{eq:clip_pred}). 
Besides, we report the upper bound~\cite{wu2019large} performance by joint training all tasks, denoted as Oracle.
All methods are based on the {\bf same pre-trained CLIP for fair comparison.}

\noindent {\bf Implementation details:} 
We deploy all methods with PyTorch~\cite{paszke2019pytorch} on Tesla V100. 
We use the \emph{same} network backbone, \ie, CLIP with ViT-B/16 for all compared methods for {\em fair comparison}. 
We experiment with two commonly used pre-trained CLIP weights, \ie, OpenAI~\cite{radford2021learning} and OpenCLIP LAION-400M~\cite{ilharco_gabriel_2021_5143773}.
The model is trained with a batch size of 64 for 5 epochs, and we use SGD with momentum for optimization. The learning rate starts from $0.001$ and decays with cosine annealing.  
Following~\cite{rebuffi2017icarl}, we use the herding~\cite{welling2009herding} algorithm to select 20 exemplars per class for rehearsal.
The context prompt length is set to 3, and the head of self-attention is set to 1. The template for classification in CLIP is kept the same as~\cite{mu2022slip}.

\begin{figure*}[t]
	\begin{center}
		\subfigure[ OpenAI weight]
		{\includegraphics[width=.65\columnwidth]{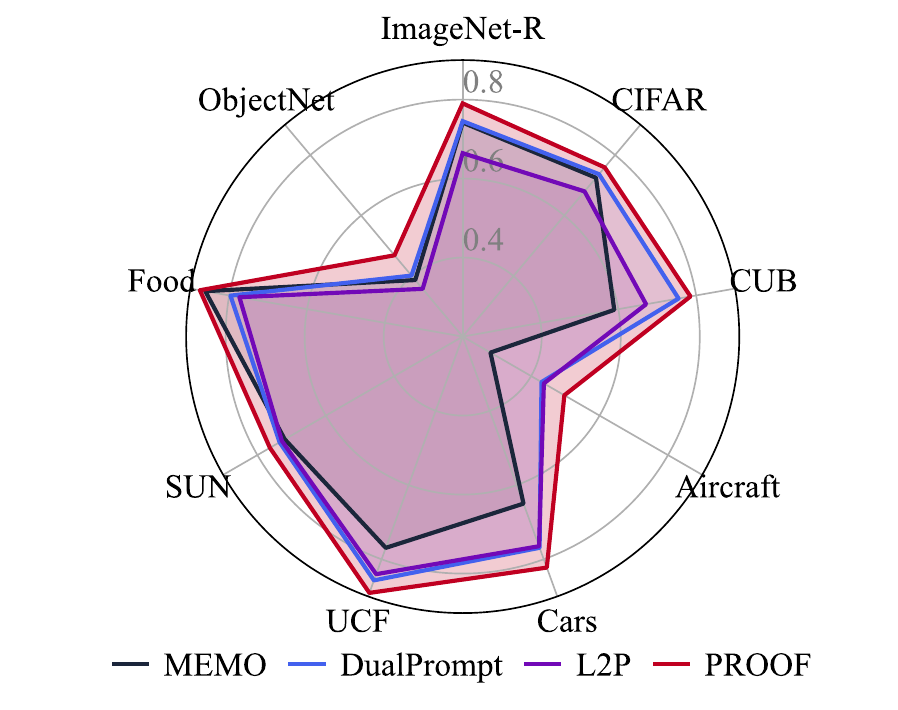}	\label{figure:ablationa}}
		\subfigure[ Compositional components]
		{\includegraphics[width=.65\columnwidth]{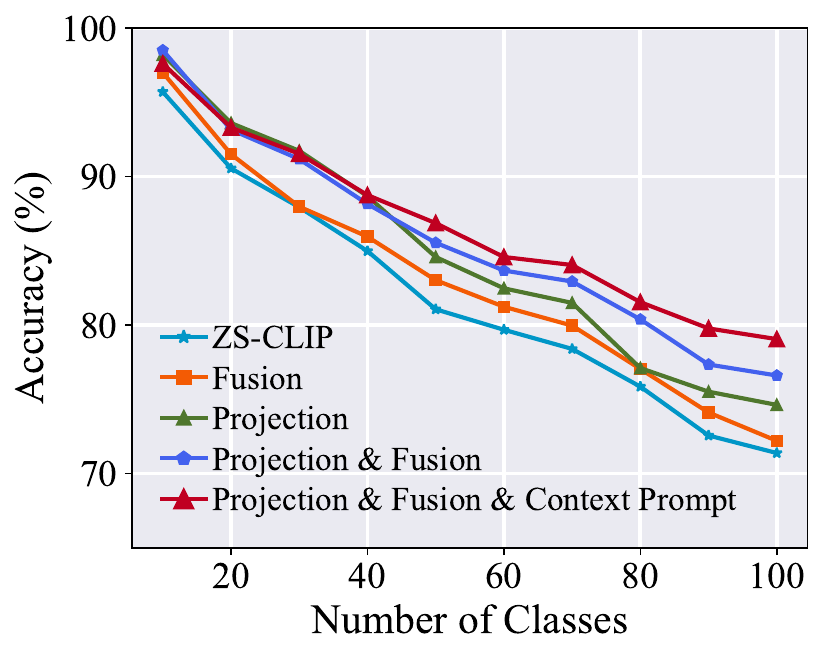}
			\label{figure:ablationb}}
		\subfigure[ Context prompt length]
		{\includegraphics[width=.65\columnwidth]{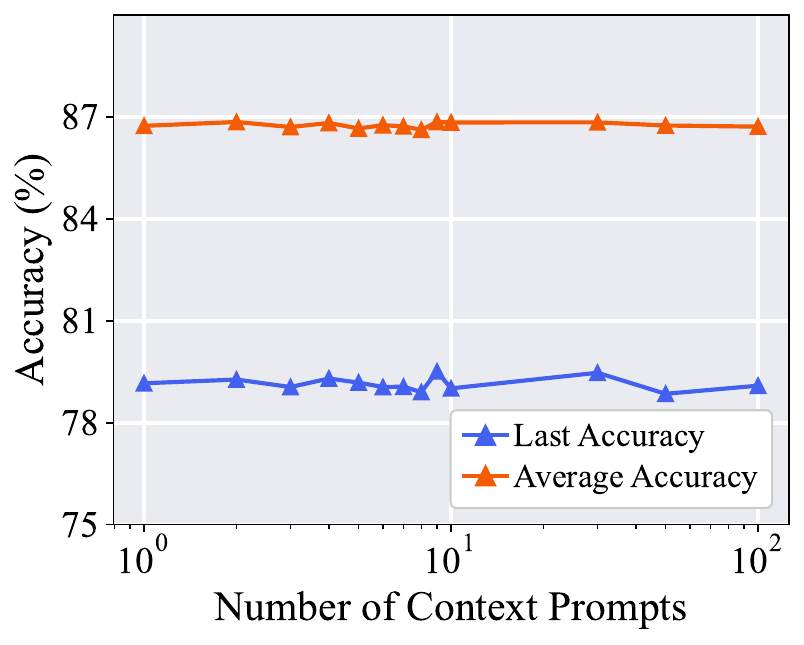}
			\label{figure:ablationc}}
	\end{center}
	\caption{Ablation study. {\bf Left:} experiments on nine benchmarks with OpenAI weights. {\bf Middle:} ablation study on compositional components in \mame. Every part improves the performance of CIL. {\bf Right:} $\mathcal{A}_B$ and $\bar{\mathcal{A}}$ with change of context prompts. The performance is robust to the change of context prompt length.
	} \label{figure:ablation}
\end{figure*}

\noindent\textbf{Evaluation Metrics:} Denote the accuracy after the $b$-th stage as $\mathcal{A}_b$, we follow~\cite{rebuffi2017icarl} to use $\mathcal{A}_B$ (last stage performance) and $\bar{\mathcal{A}}=\frac{1}{B}\sum_{b=1}^{B}\mathcal{A}_b$ (average performance) for evaluation.

\subsection{Benchmark Comparison} \label{sec:benchmark}
We report the results on nine benchmark datasets using CLIP with ViT-B/16 (OpenCLIP LAION-400M) in Table~\ref{tab:supp_benchmark} and Figure~\ref{figure:supp_benchmark}, \ref{figure:benchmark_large_base}. 
These splits include the scenarios with large and small base classes.
Notably, \name consistently achieves the best performance among all the methods compared. 
Sequential finetuning of the model with contrastive loss leads to significant forgetting, irrespective of the tuning techniques employed (\eg, LiT and CoOp). 
Since SimpleCIL and ZS-CLIP do not finetune the model parameters, they achieve competitive results by transferring the knowledge from the pre-training stage into the downstream tasks. 
However, most methods achieve better performance than ZS-CLIP, indicating the importance of incremental learning on downstream tasks.
It must be noted that the performance of L2P, DualPrompt, and CODA-Prompt are reproduced with CLIP's visual branch, which results in a different performance from the original papers.
Specifically, we can draw three key conclusions from these results. 
\begin{itemize}
	\item The first stage performance of \name surpasses that of the typical prompt learning method, CoOp, thus validating the effectiveness of learning projections for downstream tasks.
	\item The performance curve of \name consistently ranks at the top across all methods, demonstrating its capability to resist forgetting.
	\item Compared to vision-only methods (\ie, L2P, DualPrompt, CODA-Prompt, DAP), \name exhibits substantial improvement, indicating textual and visual information can be co-adapted to facilitate incremental learning. 
\end{itemize}

\subsection{Ablation Study} \label{sec:ablation}

\subsubsection{Different backbone weights}
 The comparison in Section~\ref{sec:benchmark} is based on LAION-400M pre-trained CLIP. As another popular pre-trained weight, we also explore the performance of the weights provided by OpenAI. As depicted in the figure, \name still performs the best on all datasets among all compared methods.

\subsubsection{Compositional components}
 We experiment on CIFAR100 B0 Inc10 to investigate the importance of each part in \mame. Specifically, we compare the performance of \name and its sub-modules, \ie, projections and cross-modal fusion.
The results, shown in Figure~\ref{figure:ablationb}, indicate that training expandable projections or the fusion module individually can both enhance the performance of vanilla CLIP.
This suggests that the expandable task representation and cross-modal information can help the learning process. Furthermore, when combining them together, we find `Projection \& Fusion' further show better performance than any of them, verifying that they can work together by fusing the expandable representations. Lastly, when incorporating the context prompts, the model shows the best performance among all variations, verifying the effectiveness of expandable task-specific prompts in incremental learning. Ablations verify the importance of each component in \mame.

\begin{figure*}[t]
	\begin{center}
		\subfigure[Context construction]
		{\includegraphics[width=.65\columnwidth]{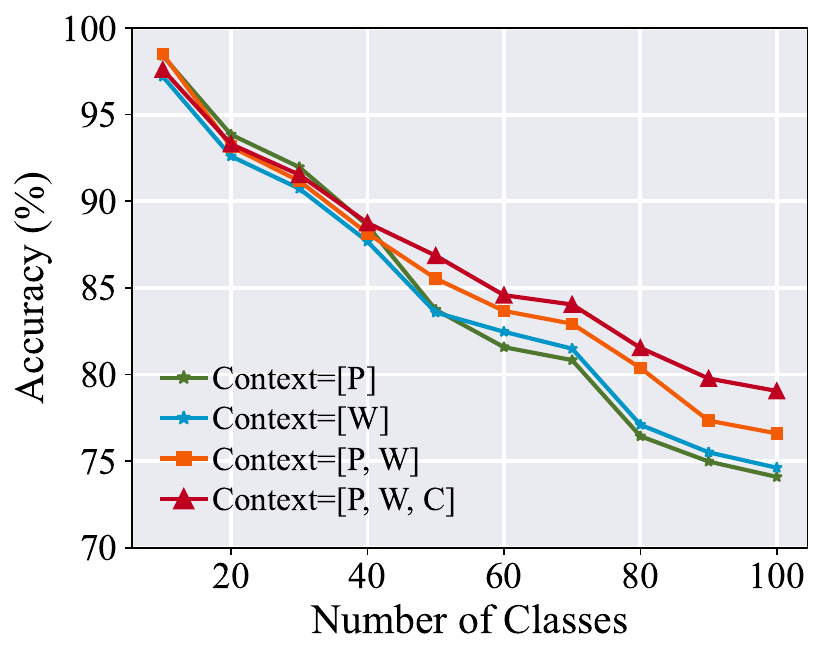}	\label{figure:ablation2a}}
		\subfigure[Projection layers]
		{\includegraphics[width=.65\columnwidth]{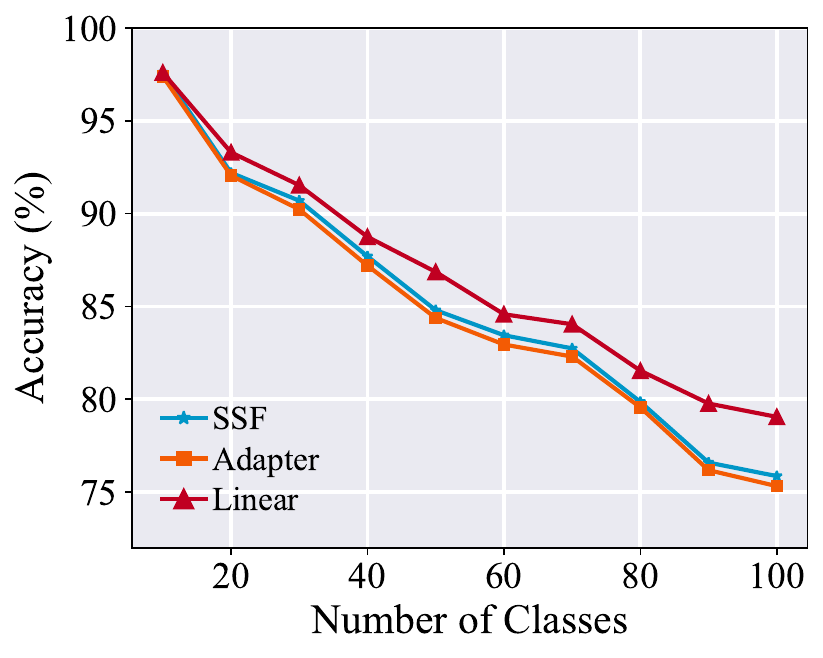}
			\label{figure:ablation2b}}
		\subfigure[Parameter size]
		{\includegraphics[width=.65\columnwidth]{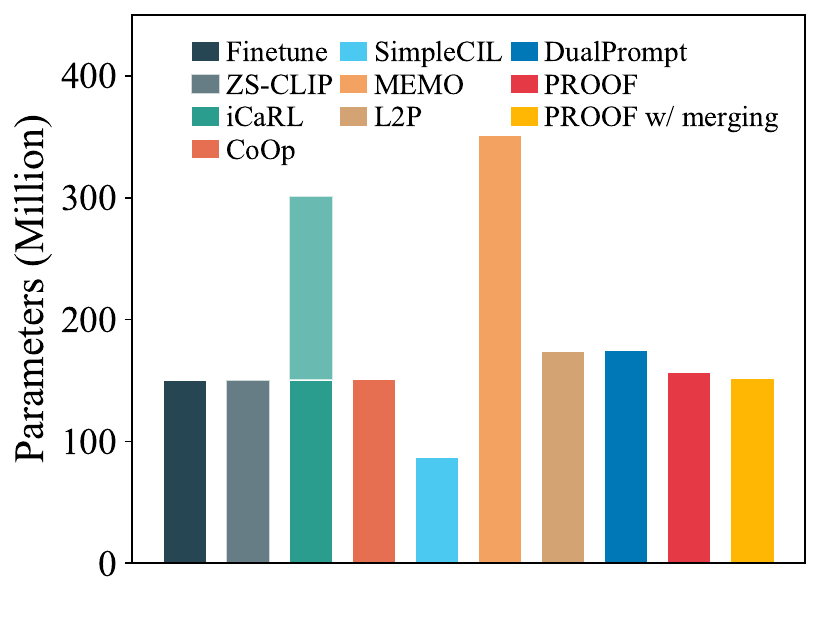}
			\label{figure:ablation2c}}
	\end{center}
	\caption{{\bf Left:} Variations of context information. The choice of using visual prototypes, textual prototypes, and context prompts as the context information achieves the best performance. {\bf Middle:} Variations of projection layers. The choice of using a single linear layer as the projection layer achieves the best performance. {\bf Right:} Number of parameters in different methods. The shaded area represents the parameters used during training but dropped during inference. 
	} \label{figure:ablation2}
\end{figure*}

\subsubsection{Number of context prompts}
 Figure~\ref{figure:ablationb} verifies the strong performance of context prompts, and we explore the appropriate length $c$ of the context prompt on CIFAR100 B0 Inc10. By varying $c$ among $\{1,2,3,4,5,6,7,8,9,10,30,50,100\}$, we report the average performance and last performance of \name in Figure~\ref{figure:ablationc}. 
 It must be noted that the prompt length $c$ is different from the task number $B$, \ie, we only change the scale of each context prompt but maintain one context prompt per incremental task. 
 Correspondingly, we find the performance of \name is robust with the change of the prompt length, indicating that context prompt only requires a small scale to encode task-specific information. 
 Hence, we set $c=3$ as the default length.

\subsubsection{Variation of context information}

 In this section, we conduct ablations to demonstrate the effectiveness of constructing $\mathbf{Context}$ with $[\mathbf{P}, \mathbf{W}, \mathbf{C}]$. Specifically, we perform experiments on CIFAR100 B0 Inc10 and change the context construction to $\mathbf{P}$ (visual prototypes only), $\mathbf{W}$ (textual prototypes only), $[\mathbf{P}, \mathbf{W}]$ (visual and textual prototypes), and $[\mathbf{P}, \mathbf{W}, \mathbf{C}]$. We keep the same classification rule for these ablations, \ie, classification via Eq.~\ref{eq:loss}. When visual/textual prototypes are not included in the context, we use the projected features without adaptation as the matching target in Eq.~\ref{eq:fuse}. The results are presented in Figure~\ref{figure:ablation2a}. We observe that using visual prototypes or textual prototypes alone yields similar performance, and the impact of adjustment is marginal. However, when both visual and textual prototypes are jointly utilized as context information, the model can learn from cross-modality and achieve better performance. Lastly, the introduction of context prompts into the context further enhances the performance of \mame, resulting in the best performance among all variations.

\subsubsection{Variation of projection types}

Apart from simple linear layers, there are other methods to implement the projection layers, such as layer-wise rescale (SSF)~\cite{lianscaling} and Adapter~\cite{houlsby2019parameter}. SSF learns a $d$-dimensional rescale parameter to project the features, while Adapter learns both the down-projection and up-projection for feature mapping. We explore the performance of these projection methods on CIFAR100 B0 Inc10 and present the results in Figure~\ref{figure:ablation2b}. The figure clearly demonstrates that using a single linear layer as the projection layer achieves the best performance among all methods, indicating its superiority. 
Furthermore, this result suggests that a simple linear mapping can effectively bridge the gap between visual and textual domains.

\begin{table}[t]
	\caption{ Experiments on by varying the number of classes per task. All methods are implemented with the same CLIP weight and the same number of exemplars. }
	\label{tab:diff-cls-order}
	\centering
	\resizebox{\columnwidth}{!}{%
		\begin{tabular}{@{}lcccccc}
			\toprule
			\multicolumn{1}{l}{\multirow{2}{*}{Method}} & 
			\multicolumn{1}{l}{\multirow{2}{*}{Exemplar}} &
			\multicolumn{2}{c}{CIFAR100} & \multicolumn{2}{c}{ImageNet-R}  \\
			&	& {$\bar{\mathcal{A}}$} & ${\mathcal{A}_B}$  
			& {$\bar{\mathcal{A}}$} & ${\mathcal{A}_B}$	\\
			\midrule
			DualPrompt~\cite{wang2022dualprompt}& \ding{51} &82.42&74.09&79.41&71.68  \\
			PLOT~\cite{chen2023plot}&\ding{51}&77.90&67.66&70.85&58.63 \\
			CODA-Prompt~\cite{smith2023coda}&\ding{51}&82.86&75.69&80.23&72.34 \\
			DAP~\cite{jung2023generating}&\ding{51}&81.12&73.56&74.38&73.68\\
			\rowcolor{LightCyan}\name & \ding{51}  & \bf 86.67& \bf 79.75& \bf83.48&\bf78.37 \\
			\bottomrule
		\end{tabular}
	}
\end{table}

\subsubsection{Parameter analysis} \label{sec:parameter}

The additional parameters in \name come from three sources: the projections, the fusion module, and the visual prototypes. 
The projection layers are implemented with a single linear layer, each containing $d\times d$ parameters, where $d$ is the embedding dimension. The cross-modal fusion is implemented with a single-head self-attention mechanism, and the number of parameters is determined by the weight matrices $W_Q$, $W_K$, and $W_V$, each containing $d\times d$ parameters. 
The visual prototypes require saving $B\times d$ features, where $B$ is the number of all classes.
The total number of extra parameters is $(2b+3)\times d^2 + B\times d$.
Hence, these extra parameters are negligible compared to the large backbone of the pre-trained CLIP model, which has approximately 150 million parameters.

\noindent\textbf{Inference Time Merging:} As defined in Eq.~\ref{eq:proj_sum}, the projected embeddings are defined as the summation of all projections. Since these projections are linear layers, we can utilize the associative law of multiplication to merge these projections:
\begin{align} \label{eq:supp_proj_sum}  
	P_i(\z)=\sum_{m=1}^b P_i^m \left(\z\right)=(\sum_{m=1}^b P_i^m)\left(\z\right)=\hat{P}_i(\z) \,.
\end{align} 
Eq.~\ref{eq:supp_proj_sum} indicates that we can merge all the projections ($P_i^1, P_i^2, \cdots, P_i^b$) into a single one ($\hat{P}_i$)  using the summation of the weights. Note that $\hat{P}_i$ has the same dimension as the single projection, which means we can alleviate the storage burden of $b$ projections into a single one. This helps us to decrease the extra parameters from $(2b+3)\times d^2 + B\times d$ to $5\times d^2 + B\times d$. Since $B$ denotes the total number of classes (which ranges from 100 to 300 in current CIL benchmarks), the second term is much smaller than the first term, and the total memory budget is limited by merging all the projections into a single one.

To provide a clear comparison of the parameter numbers for each method, we present the details in Figure~\ref{figure:ablation2c} using CIFAR100 B0 Inc10 as an example. The figure illustrates that \name has a similar parameter scale to other finetune-based methods while achieving significantly stronger performance. SimpleCIL, which only utilizes the vision branch, requires fewer parameters for the textual branch but lacks the zero-shot capability. L2P and DualPrompt also only require the vision branch but need an additional encoder to identify the appropriate prompt, resulting in a higher parameter count than \mame. Additionally, \name with projection merging further restricts the number of parameters to be similar to a zero-shot CLIP.

\subsubsection{Different number of classes per task}

In the real world, the incremental model may face a  different number of classes per task, which requires the model to tackle different numbers of classes robustly. We experiment with CIFAR100 and ImageNet-R by randomly sampling the number of classes per task, resulting in the task sequence of $\{12, 13, 10, 14,  5, 13, 13, 14, 6\}$ for CIFAR100 and $\{14, 25, 12, 17, 11, 27, 12, 12, 13, 21, 10, 16,10\}$ for ImageNet-R. The performance in Table~\ref{tab:diff-cls-order} shows that \name still shows substantial improvements in this real-world scenario.

\begin{table}[t]
	\caption{ Average and last performance of different methods on continual cross-modal retrieval tasks. The first row stands for the text retrieval task, and the second is the image retrieval task. 
	}\label{tab:supp_retrieval}
	\centering
	\begin{tabular}{@{}lccccccccccccccc}
		\toprule
		\multicolumn{1}{c}{\multirow{2}{*}{Method}}
		&
		\multicolumn{6}{c}{Image $\rightarrow$ Text }   
		\\  
		&  
		{$R_B@1$} & {$\bar{R}@1$}&	{$R_B@5$} & {$\bar{R}@5$}
		& 	{$R_B@10$} & {$\bar{R}@10$}
		\\
		\midrule
		Finetune & 48.79 & 62.89 & 76.38 & 85.04 & 85.68 & 91.84\\
		DER~\cite{yan2021dynamically} & 78.37 & 84.48 & 96.34 & 98.23 & 99.06 & 99.59 \\
		MEMO~\cite{zhou2022model} &  83.18 & 87.79 & 96.57 & 98.27 & 99.16 & 99.66 \\
		\midrule
		\rowcolor{LightCyan}	\name & \bf85.68 & \bf89.43 & \bf97.07 &\bf 98.68 & \bf99.79 &\bf 99.86 \\
	\end{tabular}
	\begin{tabular}{@{}lccccccccccccccc}
		\toprule
		\multicolumn{1}{c}{\multirow{2}{*}{Method}}
		&
		\multicolumn{6}{c}{Text $\rightarrow$ Image }   
		\\  
		&  
		{$R_B@1$} & {$\bar{R}@1$}&	{$R_B@5$} & {$\bar{R}@5$}
		& 	{$R_B@10$} & {$\bar{R}@10$}
		\\
		\midrule
		Finetune & 37.35 & 51.33 & 67.38 & 77.77 & 77.95 & 85.55\\
		DER~\cite{yan2021dynamically} & 66.71 & 74.18 & 89.63 & 93.00 & 94.84 & 96.69 \\
		MEMO~\cite{zhou2022model} & 69.53 & 76.35 & 91.89 & 94.44 & 96.09 & 97.32 \\
		\midrule
		\rowcolor{LightCyan}	\name  & \bf 72.10 & \bf78.01 & \bf93.10 &\bf 95.27 &\bf 96.92 &\bf 97.90 \\
		\bottomrule
	\end{tabular}
\end{table}

\begin{figure}[t]
	\begin{center}
		\subfigure[IR@1]
		{\includegraphics[width=.48\columnwidth]{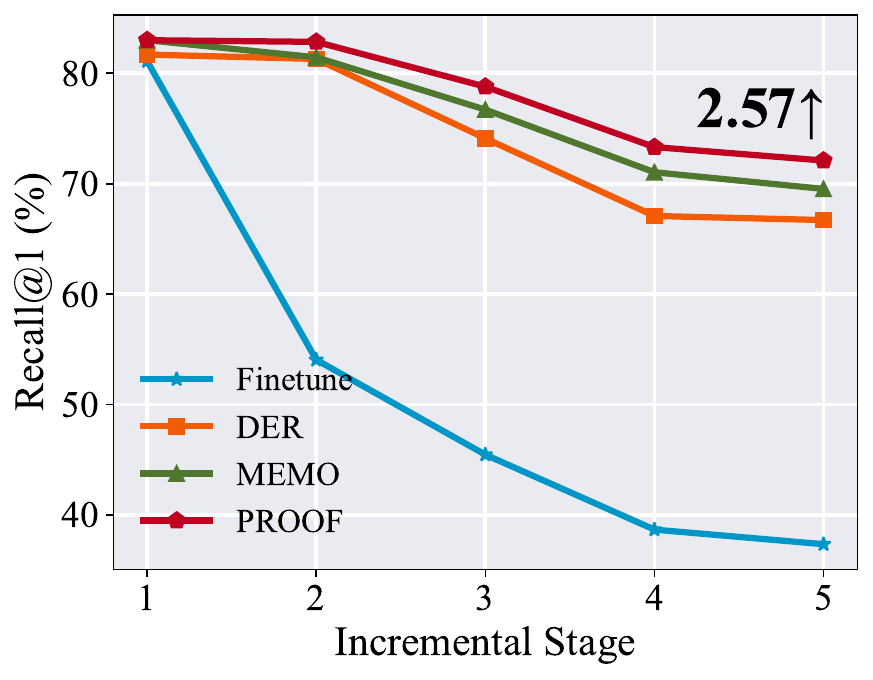}}
		\subfigure[TR@1]
		{\includegraphics[width=.48\columnwidth]{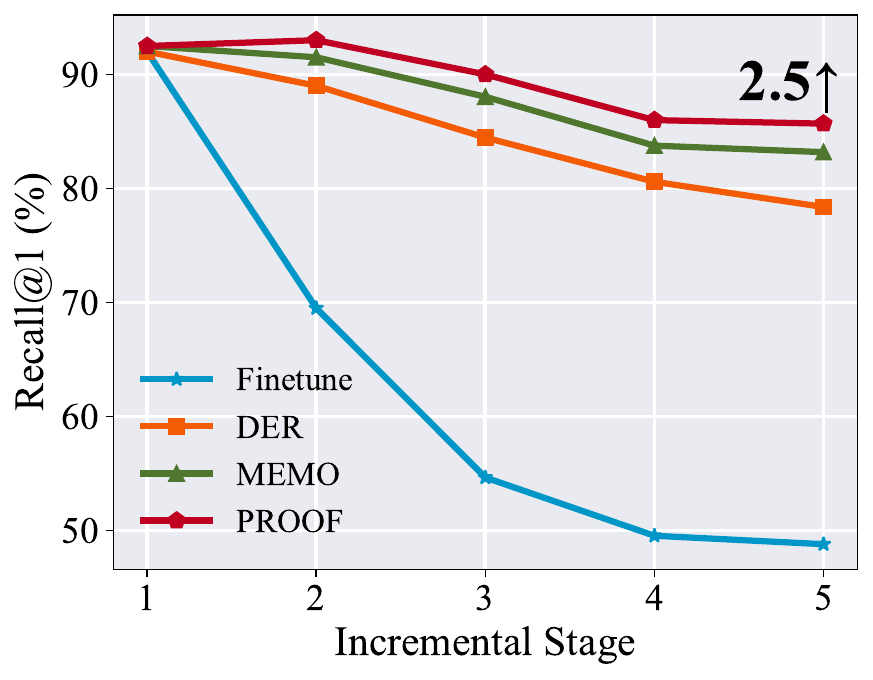}}
	\end{center}
	\caption{
		Incremental performance of each method. IR means the recall of image retrieval, and TR denotes the recall of text retrieval.
		\name consistently outperforms other compared methods with a substantial margin on the continual cross-modal retrieval task.
	}
	\label{figure:retrieval}
\end{figure}

\subsection{Extending {\scshape{Proof}} to other VLMs and other applications} \label{sec:exp:otherVLM}

In previous sections, we use CLIP as the VLM due to its popularity and representativeness. However, the field of VLM is rapidly advancing, and various models are available. In this section, we extend \name to another widely used VLM, \ie, BEiT-3~\cite{wang2022image}, focusing on the cross-modal retrieval task.
BEiT-3 is a popular VLM that demonstrates promising performance across multiple vision-language tasks. When finetuning BEiT-3 for cross-modal retrieval, it functions as a {\em dual encoder}, similar to CLIP, featuring a dual-branch structure. As the retrieval task differs from classification, we adopt a degradation of \name by solely employing the projection expansion strategy without implementing cross-modal fusion.

For evaluation, we employ the Flickr30K dataset~\cite{plummer2015flickr30k} to assess the performance of incremental cross-modal retrieval. Flickr30K comprises 31,783 images collected from the Flickr image-sharing platform, encompassing diverse themes such as daily life, travel, people, food, and scenes. Each image in the dataset is accompanied by {\em five} manually annotated textual descriptions, which provide descriptive information capturing the main content and context of the images. To formulate an incremental data stream, we utilize keyword matching to identify images containing different actions (\eg, walk, stand, run, ride, play). Then, we split the training instances into five subsets based on these specific actions. 
To create a balanced testing set, we maintain a 5:1 training-to-testing ratio for splitting the training and testing pairs. 

We employ standard cross-modal retrieval measures  for evaluation, namely $R@1$, $R@5$, and $R@10$. The retrieval is conducted in two directions: image $\rightarrow$ text and text $\rightarrow$ image. Similarly to the CIL evaluation, we report the last recall $R_B@1$ and the average recall $\bar{R}@1$ across incremental stages.
To provide a comparative analysis, we compare \name against typical finetuning as the baseline and modify MEMO~\cite{zhou2022model} and DER~\cite{yan2021dynamically} for comparison. These methods represent state-of-the-art CIL approaches that can be adapted with minor modifications to the current task. However, methods such as L2P and DualPrompt are unsuitable for cross-modal retrieval tasks as they do not focus on cross-modal matching.

The experimental results are presented in Table~\ref{tab:supp_retrieval}, and Figure~\ref{figure:retrieval}. As evident from these figures, finetuning the model with new concepts leads to catastrophic forgetting in cross-modal retrieval tasks. However, equipping the model with incremental learning abilities alleviates forgetting. Among all the compared methods, \name consistently achieves the best performance across different retrieval tasks and metrics, thereby verifying its effectiveness in mitigating forgetting in VLMs.
In summary, \name performs competitively against other algorithms even with different VLMs and continual learning settings.

\begin{figure}[t]
	\begin{center}
		\subfigure[ TV100 Base0 Inc10]
		{\includegraphics[width=.48\columnwidth]{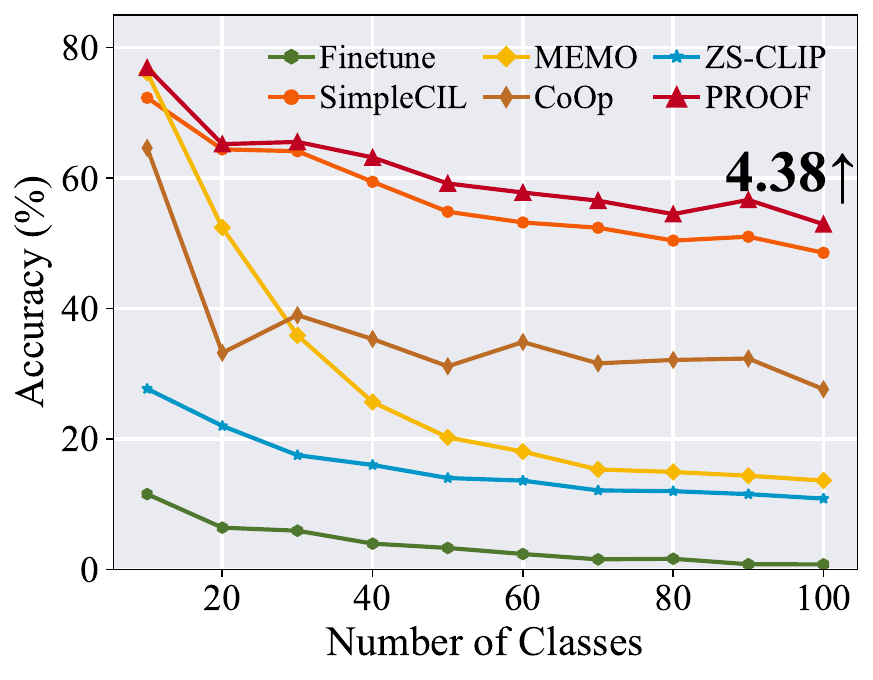}}
		\subfigure[ TV100 Base50 Inc10]
		{\includegraphics[width=.48\columnwidth]{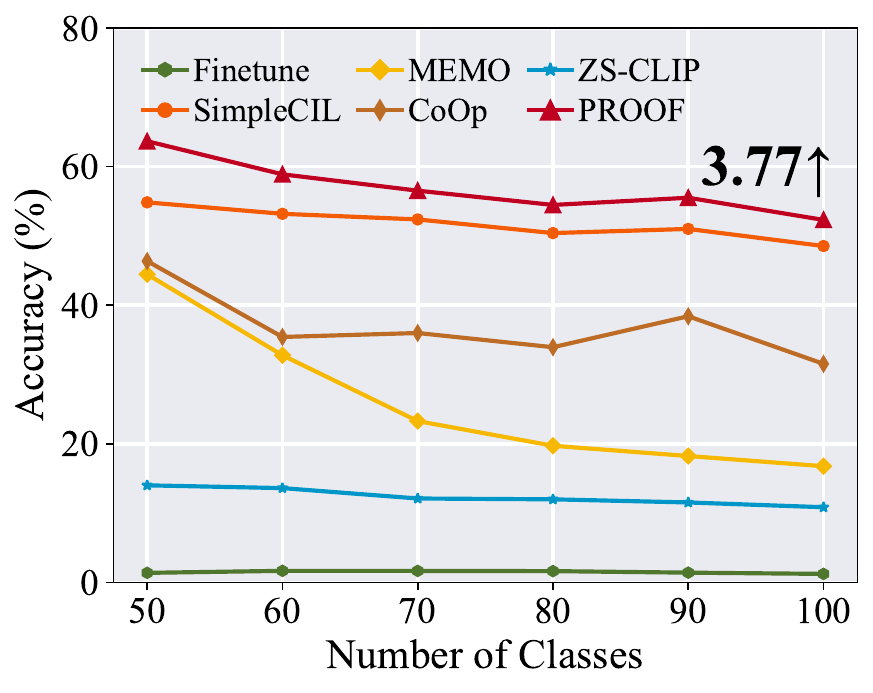}}
	\end{center}
	\caption{
		Experiments on TV100, a non-overlapping dataset containing images of TV series after 2021. \name outperforms other compared methods by a substantial margin.
	}
	\label{figure:supp_tv100}
\end{figure}

\begin{figure*}[t]
	\begin{center}
		\subfigure[ Unseen class accuracy]
		{\includegraphics[width=.65\columnwidth]{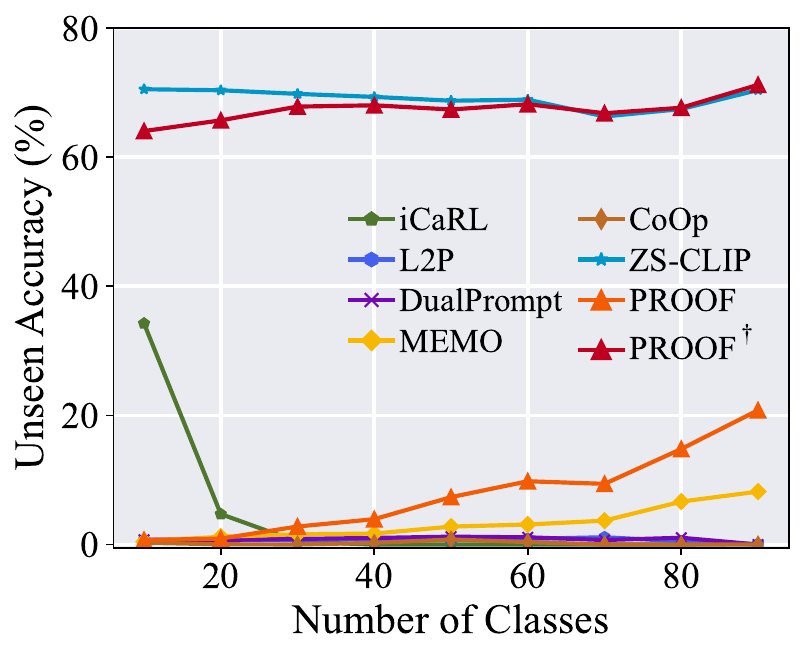}
			\label{figure:zs-a}}
		\subfigure[ LAION score]
		{\includegraphics[width=.65\columnwidth]{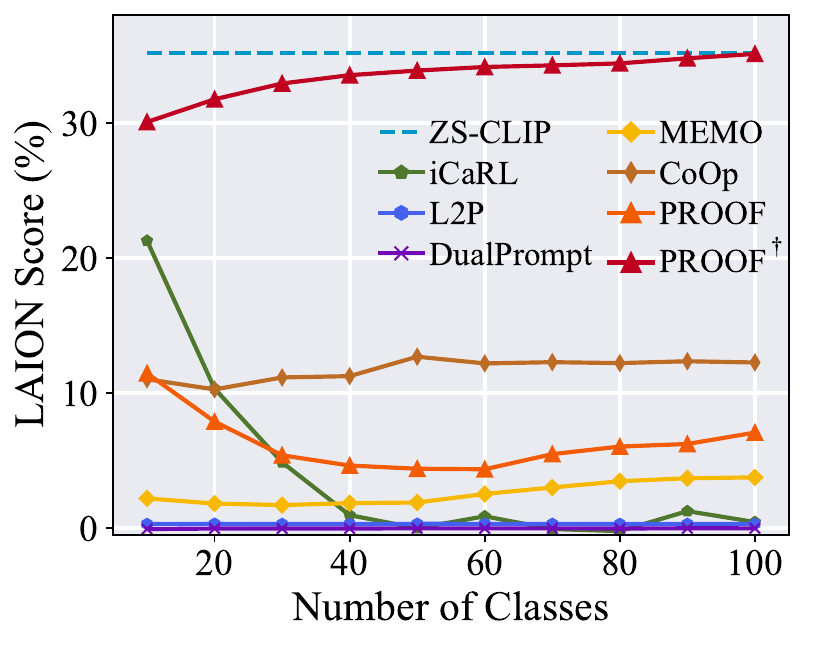}
			\label{figure:zs-b}}
		\subfigure[ $\mathcal{A}_\text{S} \,,\mathcal{A}_\text{U} \,, \mathcal{A}_\text{HM}$  ]
		{\includegraphics[width=.65\columnwidth]{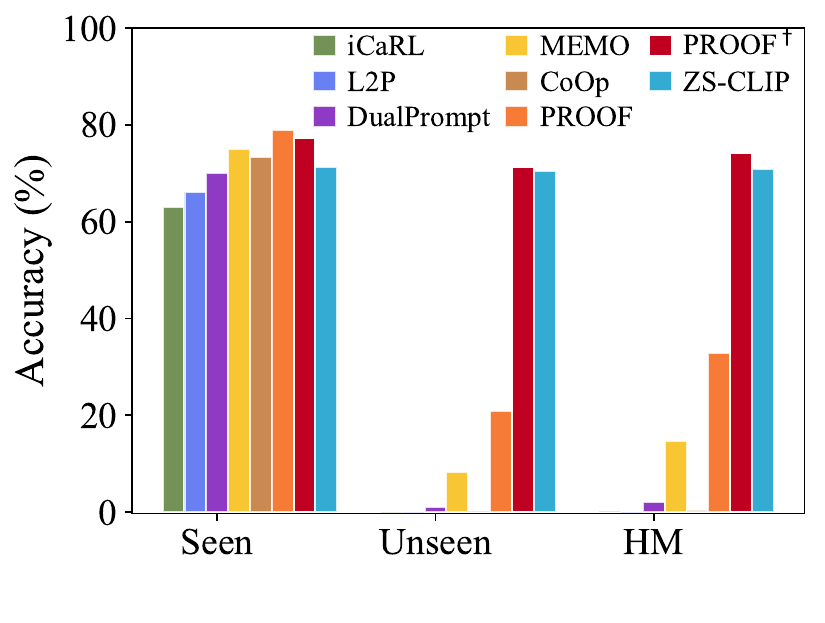}
			\label{figure:zs-c}}
	\end{center}
	\caption{ Experiment on zero-shot performance. {\bf Left:} accuracy on unseen classes during incremental learning. {\bf Middle:} LAION score during incremental learning. {\bf Right:} accuracy of seen, unseen, and harmonic mean (HM) at the last incremental stage.  \mame$^\dagger$ strikes a balance between adaptivity and the ZS performance.
	} \label{figure:zs}
\end{figure*}

\subsection{CIL with non-overlapping dataset}
 We have verified \mame's performance on benchmark CIL datasets in Section~\ref{sec:benchmark}. However, one may argue that these benchmark datasets may have data overlapping with CLIP's pre-training dataset. Hence, we manually collect a new dataset for TV series classification with TV series after the publication of CLIP, namely TV100, for evaluation.

{\noindent\textbf{Dataset Construction:}} CLIP is proposed in 2021, which is trained with image-text pairs (before the year 2021) collected from the Internet. Hence, if we can collect a new dataset after 2021,  we can tell that {\em CLIP does not know the new knowledge.} To achieve this goal, we select a field with new classes emerging every day, \ie, the TV series. Specifically, we manually search for TV series from IMDB and collect the items released after 2021\footnote{For those series with multiple seasons, we directly drop them since CLIP may have seen a former season, \eg, Stranger Things Season 4 is released in 2022, while Stranger Things Seasons 1, 2, and 3 are released before 2021.}. Afterward, we download the related images on Google by searching the keyword ``{\bf [NAME] TV Series},'' where [NAME] is the name of the TV series. The downloaded images are then processed manually to delete repeated and meaningless ones. Hence, we can get a large dataset that contains around 800 classes.

However, some of these classes may not be ``new'' for a pre-trained CLIP, \eg, ``The Kardashians'' was released in 2022 while it is not a new concept for CLIP because the Kardashian–Jenner family has been popular in America since the last century. A similar phenomenon also occurs in ``The Snoopy Show'' (Snoopy is a famous cartoon character) and ``The Cuphead Show'' (Cuphead is a video game released in 2017). Hence, we need to select some challenging  classes that CLIP does not know from the TV series pool. Correspondingly, we use a pre-trained CLIP to rank the difficulty of these classes by measuring the zero-shot accuracy of each image and the text ``a photo of the TV series [CLASS].'' We choose the top-100 hard classes based on the zero-shot accuracy and construct the TV100 dataset. 
Surprisingly, a pre-trained CLIP only achieves around 10\% accuracy on this dataset, verifying that CLIP does not master these classes. Besides, since the dataset is collected after the publication of CLIP,  there is no class overlapping between pre-trained CLIP and TV100. 

Correspondingly, we conduct experiments on this new dataset. With the other settings the same as the main paper, we select two dataset splits (\ie, Base0 Inc10 and Base50 Inc10) and report the results in Figure~\ref{figure:supp_tv100}. We can summarize two main conclusions from the figure. Firstly, zero-shot CLIP performs poorly on this dataset, verifying that this dataset perfectly serves as the benchmark to evaluate the continual learning ability of pre-trained CLIP. Secondly, \name still outperforms other competitors by a substantial margin, verifying its strong performance in real-world continual learning tasks.

\subsection{Exploring Zero-Shot Performance}

CLIP is known to have the zero-shot (ZS) ability, \ie, even if the model has not been trained for recognizing the image, it can still predict the possibility of an image $\x$ belonging to the class $y$ by matching the cosine similarity via Eq.~\ref{eq:clip_pred}.
The strong generalizability of CLIP makes it a popular model in computer vision.
However, in CIL, the model is {\em continuously} updated with the downstream task, which weakens the generalizability and harms the ZS performance~\cite{wortsman2022robust} on subsequent tasks. In this section, we explore the zero-shot performance degradation of CLIP and propose a variation of \name to maintain the zero-shot performance.

\noindent\textbf{Evaluation protocol for ZS performance:} Current CIL methods focus on evaluating `seen' classes, \ie, evaluating $\mathcal{Y}_b=Y_1 \cup \cdots Y_b$ after learning task $b$. However, since CLIP exhibits ZS performance, we can also assess the performance on `unseen' classes $\mathcal{Y}_u=Y_{b+1} \cup \cdots Y_B$ to investigate the ZS performance. Correspondingly, we can obtain the performance metrics $\mathcal{A}_\text{S}$ (seen classes), $\mathcal{A}_\text{U}$ (unseen classes), and $\mathcal{A}_\text{HM}$ (harmonic mean of $\mathcal{A}_{S}$ and $\mathcal{A}_{U}$) after each task. 
Additionally, based on the LAION-400M~\cite{schuhmann2021laion} pre-trained CLIP, we also utilize a subset of 10,000 image-text pairs from LAION-400M, and calculate the matching score of them, \ie, cosine similarity of image-text embeddings. 
We denote the average matching score as {\em LAION score}, which indicates the matching degree of the adapted model on the {\em upstream} tasks. 
Given the relationship between generalizability and the upstream task, the LAION score serves as an effective measure of ZS performance.

\noindent\textbf{Results:} We compare the aforementioned measures on CIFAR100 B0 Inc10. Apart from the compared methods in Section~\ref{sec:benchmark}, we also report a variation of \mame, namely \mame$^\dagger$. 
The only difference lies in the design of the projection, where \mame$^\dagger$ uses a {\em residual} format as the output:
\begin{equation}  \textstyle \notag
	P_i(\z)=\sum_{m=1}^b \left(P_i^m \left(\z\right) + \z\right), \quad  P_t(\w)=\sum_{n=1}^b \left(P_t^n \left(\w\right) + \w\right)  \,.
\end{equation}
To investigate the ZS performance as model updates, we show the accuracy on unseen classes $\mathcal{A}_\text{U}$ along incremental stages in Figure~\ref{figure:zs-a}, where ZS-CLIP shows the best performance. 
Correspondingly, due to the incorporation of pre-trained information (\ie, $\z$ and $\w$) into the projected features, \mame$^\dagger$ maintains competitive ZS performance. 
It indicates that reflecting pre-trained information helps to maintain generalizability.
Conversely, other methods experience a decline in ZS performance as their focus shifts to downstream tasks.
We observe a similar trend in Figure~\ref{figure:zs-b}, where \mame$^\dagger$ achieves a LAION score similar to that of ZS-CLIP. 
Lastly, we report $\mathcal{A}_\text{S},\mathcal{A}_\text{U}, \mathcal{A}_\text{HM}$ in the last incremental stage in Figure~\ref{figure:zs-c}.
We can infer a {\em trade-off} between the adaptivity on downstream tasks and the generalizability of ZS performance. Compared to \mame, \mame$^\dagger$ sacrifices the adaptivity to maintain ZS performance, striking a balance between seen and unseen classes. Therefore, when ZS performance is essential, using \mame$^\dagger$ is the preferred choice.

\section{Conclusion} \label{sec:conclusion}

Real-world learning systems necessitate the ability to continually acquire new knowledge. In this paper, we aim to equip the popular VLM with the CIL ability. Specifically, we learn the expandable projections so that visual and textual information can be aligned incrementally. 
This expansion technique allows for integrating new concepts without compromising previous ones.
Additionally, we enforce cross-modality fusion with the self-attention mechanism, where visual and textual information are jointly adapted to produce instance-specific embeddings. 
Extensive experiments validate the effectiveness of our proposed \name in various VLMs and various continual learning scenarios. We also demonstrate that a simple variation can preserve the model's zero-shot capability. Future work includes extending the model to exemplar-free scenarios.

\ifCLASSOPTIONcompsoc

\bibliographystyle{unsrt}
\bibliography{paper}

\ifCLASSOPTIONcaptionsoff
  \newpage
\fi

\end{document}